\documentclass[preprint,authoryear,12pt]{elsarticle} 

\usepackage{graphics}
\usepackage{graphics}

\usepackage{float}
\usepackage{graphicx}
\usepackage{wrapfig}

\usepackage[labelfont=bf]{caption}
\usepackage{subcaption}
\usepackage{color}
\usepackage{amsmath} 
\usepackage{mathtools} 
\usepackage{amsfonts} 
\usepackage{amssymb}

\usepackage{array}

\usepackage{booktabs}
\usepackage{rotating}
\usepackage{multirow}
\usepackage{multicol}
\usepackage{lscape}

\usepackage{xcolor}
\usepackage{xargs}                      
\usepackage{calc}

\usepackage{url}

\usepackage[export]{adjustbox}

\usepackage{color}
\usepackage{colortbl}

\definecolor{babyblue}{rgb}{0.54, 0.81, 0.94}
\definecolor{armygreen}{rgb}{0.29, 0.33, 0.13}
\definecolor{brightlavender}{rgb}{0.75, 0.58, 0.89}
\definecolor{aqua}{rgb}{0.0, 1.0, 1.0}
\definecolor{caribbeangreen}{rgb}{0.0, 0.8, 0.6}
\definecolor{reddish}{rgb}{0.82, 0.1, 0.26}
\definecolor{caribbeangreen}{rgb}{0.31, 0.78, 0.47}
\definecolor{jasper}{rgb}{0.84, 0.23, 0.24}
\definecolor{red}{rgb}{1.0, 0.0, 0.0}
\definecolor{green}{rgb}{0.0, 1.0, 0.0}
\definecolor{blue}{rgb}{0.0, 0.0, 1.0}
\definecolor{darkgreen}{rgb}{0.1, 0.7, 0.1}
\definecolor{darkblue}{rgb}{0.1, 0.1, 0.7}
\definecolor{red}{rgb}{0.7, 0.1, 0.1}
\definecolor{coral}{rgb}{1.0, 0.5, 0.31}

\usepackage[colorlinks=true,linkcolor=coral,citecolor=coral]{hyperref}

\definecolor{darkgreen}{rgb}{0.53, 0.66, 0.42}

\usepackage[ruled,vlined]{algorithm2e}

\journal{Journal}

\begin{document}

\begin{frontmatter}




\title{Machine Learning Methods for Brain Network Classification: Application to Autism Diagnosis using Cortical Morphological Networks}


\author{Ismail Bilgen\corref{same}  \fnref{BASIRA}}
\author{Goktug Guvercin\corref{same} \fnref{BASIRA}}
\author{Islem Rekik \corref{cor} \fnref{BASIRA,DUNDEE}}

\address[BASIRA]{BASIRA lab, Faculty of Computer and Informatics, Istanbul Technical University, Istanbul, Turkey}
\address[DUNDEE]{School of Science and Engineering, Computing, University of Dundee, UK}

\cortext[same]{Co-first authors: ibilgen@itu.edu.tr and guvercing@itu.edu.tr}

\cortext[cor]{Corresponding author; Dr Islem Rekik (irekik@itu.edu.tr), \url{http://basira-lab.com/}, GitHub code: \url{https://github.com/basiralab/BrainNet-ML-ToolBox}} 




\begin{abstract}
	
\textbf{-Background.} Autism spectrum disorder (ASD) affects the brain connectivity at different levels. Nonetheless, non-invasively distinguishing such effects using magnetic resonance imaging (MRI) remains very challenging to machine learning diagnostic frameworks due to ASD heterogeneity. So far, existing network neuroscience works mainly focused on functional (derived from functional MRI) and structural (derived from diffusion MRI) brain connectivity, which might not capture relational morphological changes between brain regions. Indeed, machine learning (ML) studies for ASD diagnosis using morphological brain networks derived from conventional T1-weighted MRI are very scarce.

\textbf{-New Method.} To fill this gap, we leverage crowdsourcing by organizing a Kaggle competition to build a pool of machine learning pipelines for neurological disorder diagnosis with application to ASD diagnosis using cortical morphological networks derived from T1-weighted MRI.

\textbf{-Results.} During the competition, participants were provided with a training dataset and only allowed to check their performance on a public test data. The final evaluation was performed on both public and hidden test datasets based on accuracy, sensitivity, and specificity metrics. Teams were ranked using each performance metric separately and the final ranking was determined based on the mean of all rankings. The first-ranked team achieved 70\% accuracy,  72.5\% sensitivity, and 67.5\% specificity, while the second-ranked team achieved 63.8\%, 62.5\%, 65\% respectively.

\textbf{-Conclusion.} Leveraging participants to design ML diagnostic methods within a competitive machine learning setting has allowed the exploration and benchmarking of wide spectrum of ML methods for ASD diagnosis using cortical morphological networks. 

\end{abstract}

\begin{keyword}
	Neurological disorders, Machine Learning, Computer-Aided Diagnosis, A Python Toolbox for Network Classification, Autism Spectrum Disorder, 
\end{keyword}

\end{frontmatter}

\section{Introduction}


Autism spectrum disorder (ASD) is a neuropsychiatric condition that impairs behavioral and cognitive functions of children such as communication and social interaction. The main symptoms of ASD are restricted and repetitive behaviors and interests. The number of ASD cases are increasing in the world over time as reported in \citep{baio2018}. 
The symptoms of ASD generally appear in the first two years and tend to be long-life persistent. Nevertheless, timely treatments can improve the symptoms and abilities to function substantially. Therefore, the early accurate diagnosis of ASD is crucial to develop specialized interventions \citep{zwaigenbaum2015}. However, the diagnosis of ASD is very challenging due to its complex nature and highly heterogeneous symptoms \citep{zhao2018}. 

Several studies in neuroimaging using different non-invasive brain imaging modalities such as functional MRI (fMRI) and diffusion MRI (dMRI) were proposed to overcome this challenge \citep{zhao2018, anderson2011, eslami2019, dekhil2019, heinsfeld2018, brown2016}. Although such studies advanced our understanding of brain changes in ASD subjects on functional and structural connectivity levels, they overlooked relational morphological changes between brain regions. To address this gap in network neuroscience \citep{Fornito:2015,Bassett:2017}, a few recent studies investigated the potential of  cortical morphological networks (CMNs), derived solely from T1-weighted MRI in distinguishing between the autistic and typical cortices \citep{soussia2017,soussia2018, morris2017,georges2020}. Notably, several works investigated the change in morphology at a brain region level \citep{postema2019, itahashi2015, yang2016}, however, these did not investigate the changes in one brain region of interest (ROI) \emph{in relation} to another ROI. On the other hand, such morphological relationship between pairs of ROIs can be nicely modeled using morphological brain networks, where the morphological connectivity between two regions encodes their dissimilarity in morphology as introduced in \citep{mahjoub2018brain}.

Although these few seminal works were the first to investigate how ASD affects CMNs \citep{soussia2017,soussia2018, morris2017,georges2020}, they were based on particular machine learning (ML) methods, which leaves us with a wider sprectrum of rich and diverse ML methods that are fully unexplored for ASD diagnosis. On the other hand, crowdsourcing has emerged as a framework to address computational challenges in many areas such as biomedical and genomics \citep{rodriguez2016, belcastro2018, marbach2012wisdom}, which accelerates exploring and benchmarking both existing and novel approaches, and improves the robustness of solutions. For this purpose, we organized an in-class challenge via \cite{Kaggle}\footnote{\url{https://www.kaggle.com/}}, where participants aim to classify ASD/NC subjects using solely CMNs derived from maximum principal curvature of the cortical surface. Teams were ranked based on three classification performance metrics: accuracy, sensitivity and specificity, evaluated on a a hidden test dataset. The final ranks of teams were determined by summing ranks on each individual metric. This challenge fills the gap emerging from the lack of studies on morphological brain networks for ASD diagnosis, by enabling an assessment of a wide range of methods through standardized performance metrics.



In this manuscript, we present the results for the competition and describe the computational approaches of the top 20 ranked teams. We provide the comparison and comprehensive characterization of different ASD/NC ML classification methods on cortical morphological networks and interpret the performance of each method. Promoting open ML in network neuroscience, we have shared the Python network classification codes by the top 20 participating teams on BASIRA Lab GitHub: \url{https://github.com/basiralab/BrainNet-ML-ToolBox}, which were polished by the second-first author.

\section{Methods}

\subsection{Competition Organization}

The diagnosis of neurological disorders, such as ASD, can be formulated as a classification task in machine learning. In this case, the learning model being chosen for this task is fed by a training dataset in which each sample is represented with brain connectome map (e.g., CMN) of a subject and a binary classification label of the subject as with or without disorder. In the case of highly-qualified generalizability of the ML model to unseen data in the testing phase, such a learning model can be a pre-eminent method for the diagnosis of a particular brain disorder.  To evaluate the generalizability of a wide spectrum of ML methods in diagnosing ASD patients using CMNs, a competition was set up in Kaggle platform. The teams participating in this competition were provided with two datasets for training and testing phases, respectively. After training their designed ML frameworks, the teams were allowed to produce a prediction label list for the test samples, and to submit those predictions to the competition platform under a limited number of rights so as to check the accuracy of their models. In this way, they could change their learning algorithm or make some modifications on it like parameter tuning or utilizing some preprocessing techniques.

At the end of the competition, the teams were requested to propose their default ML frameworks so that they could be evaluated on a testing set in terms of accuracy, sensitivity, and specificity. Based on each evaluation metric, the teams were ranked. Ultimately, the overall competition rank was defined by the summation of the three metric ranks of each team.


\subsection{Analysis of Machine Learning Methods in the Brain Network Classification Kaggle Competition}

\begin{figure}[h!]
\centering
  \includegraphics[width=1\textwidth]{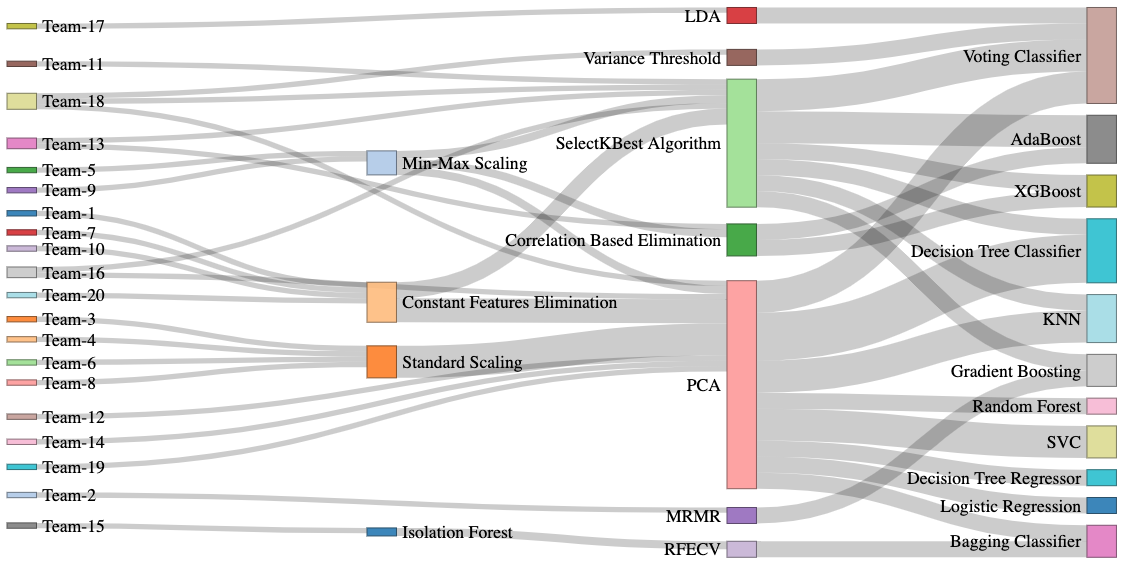}
  \caption{Trends of machine learning methods for brain network classification in Kaggle competition. The thickness of the link between two objects indicates the relative quantity of usage from left to right.}
  \label{fig:trend}
\end{figure}

In this section, we provide an overview of the machine learning pipelines that have been proposed by the top 20 leading teams in the competition. All methods laying the foundation of those pipelines are examined under three major ML categories: (1) preprocessing techniques, (2) dimensionality reduction methods, and, (3) learning models (Figure~\ref{fig:trend}).

\subsubsection{Preprocessing Techniques}

Preprocessing techniques are the algorithms on the front line of a machine learning pipeline composed of several steps such as data preparation, dimension reduction, model training, validation, and testing. The main reason why they are commonly adopted is that they can improve poor-quality data in some aspects such as outlier detection and feature scaling as well as imputation, thereby preparing a more polished data for further steps of learning process.

\begin{figure}[h!]
\centering
  \includegraphics[width=1\textwidth]{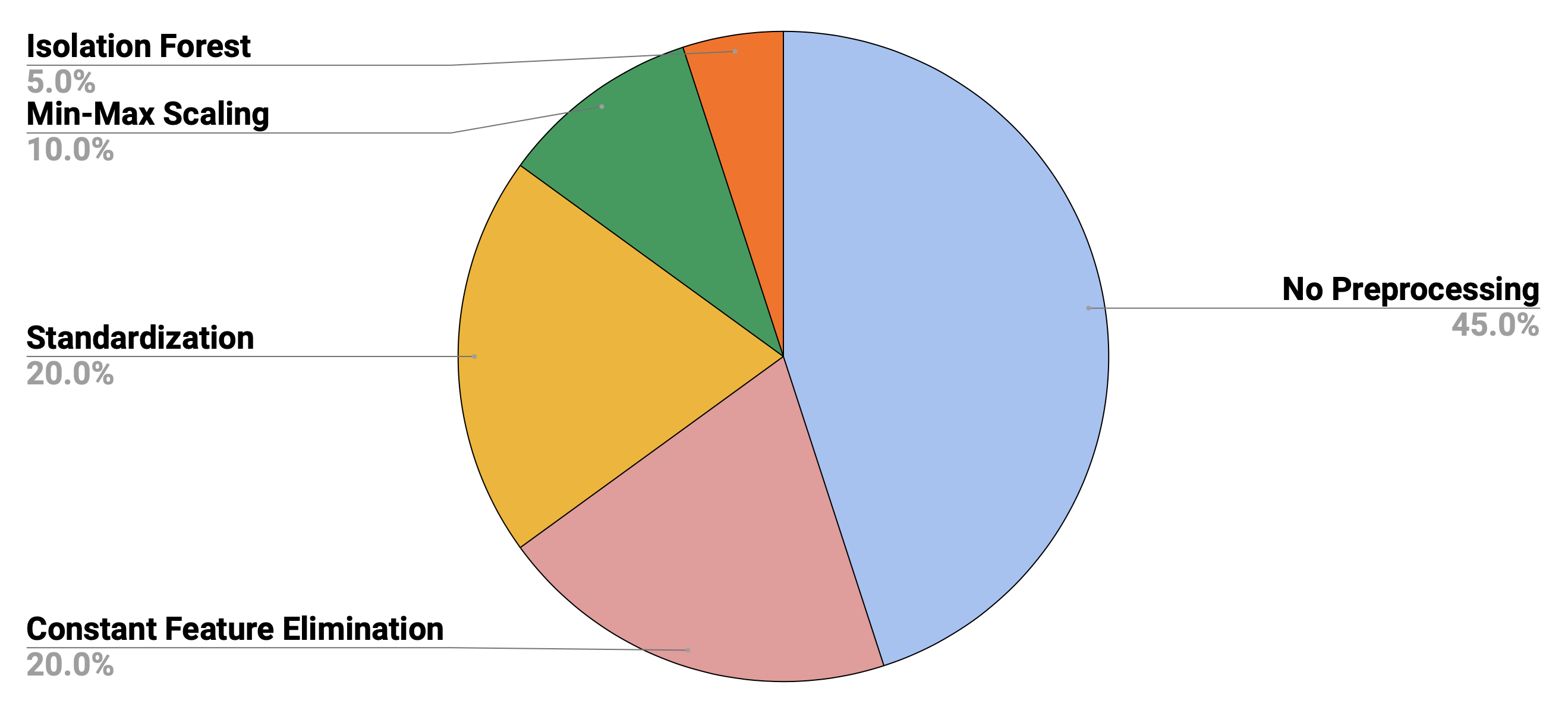}
  \caption{Brain network data preprocessing techniques used  in  the  brain network classification Kaggle competition for distinguishing between autistic and healthy subjects.}
  \label{fig:preprocessing}
\end{figure}

In the competition, there are totally four preprocessing techniques that have been deployed in machine learning pipelines proposed by the participating teams, which are illustrated in Figure \ref{fig:preprocessing}. Almost half of the teams have not preferred to perform any data preprocessing. On the other hand, feature scaling techniques (standardization and min-max scaling) have accounted for approximately one-third of total usage. While the elimination of constant features in the dataset has been leveraged by 4 teams, solely 1 team has utilized Isolation Forest algorithm for preprocessing. 

The features of the datasets used in some machine learning tasks generally vary in range, and this issue decreases the performance of many learning models and dimensionality reduction methods considerably \citep{géron2017hands-on}. For instance, gradient descent algorithm cannot modify the weights of the features with different scales at equal rate, which impacts all learning algorithms utilizing gradient descent such as logistic regression and support vector machines (SVM). In addition to this, the dimensionality reduction methods that shrink the feature space depending on the variance of the features including Principal Component Analysis (PCA) and Linear Discriminant Analysis (LDA) can be misled by large variability in feature scales. Standardization, also known as standard scaling, and min-max scaling aim to adjust all features to same scale. While standardization shifts the distribution of values that a particular feature takes over the whole dataset to zero mean and assigns it a unit standard deviation by z-score calculation, min-max scaling tries to accumulate the values of features in 0-1 range \citep{raschka2014feature}.

Apart from those feature scaling operations (standardization and min-max scaling), some competitors have also opted for eliminating six different constant features. 
The main reason why such  operation has been put into practice is that unchanged features do not have any effect on the variation of ground truth labels, thereby providing no contribution to the target classification task. Since it decreases the number of features in the dataset, it can also be regarded as a dimensionality reduction technique rather than a preprocessing operation. 

The data samples of a particular dataset  generally live in a high dimensional space where they follow a specific pattern or trend, which is the general principle that learning models aim to learn to make generalization to testing datasets (i.e., new unseen samples). However, some of those samples may have different characteristics or behavior that do not adhere to that trend properly, which are called as outliers. Isolation forest algorithm aims to identify and isolate those outliers, also called as anomalies, by splitting the feature space in which the samples are nested. 
This is based on a recursive feature space partitioning process where outlier samples become isolated faster than normal \citep{liu2008isolation}.

\subsubsection{Dimensionality Reduction Methods}

Dimensionality reduction methods used by the teams in this competition have been significantly shaped by some fundamental ML issues, one of which is the curse of dimensionality. Since each sample of the dataset is nested in high dimensional space due to the large  number of morphological features (i.e., connectivity weights) derived from the CMN, several designed learning models leveraged dimensionality reduction techniques to shrink the feature space prior to the training phase. 

\begin{figure}[h!]
\centering
  \includegraphics[width=0.8\textwidth]{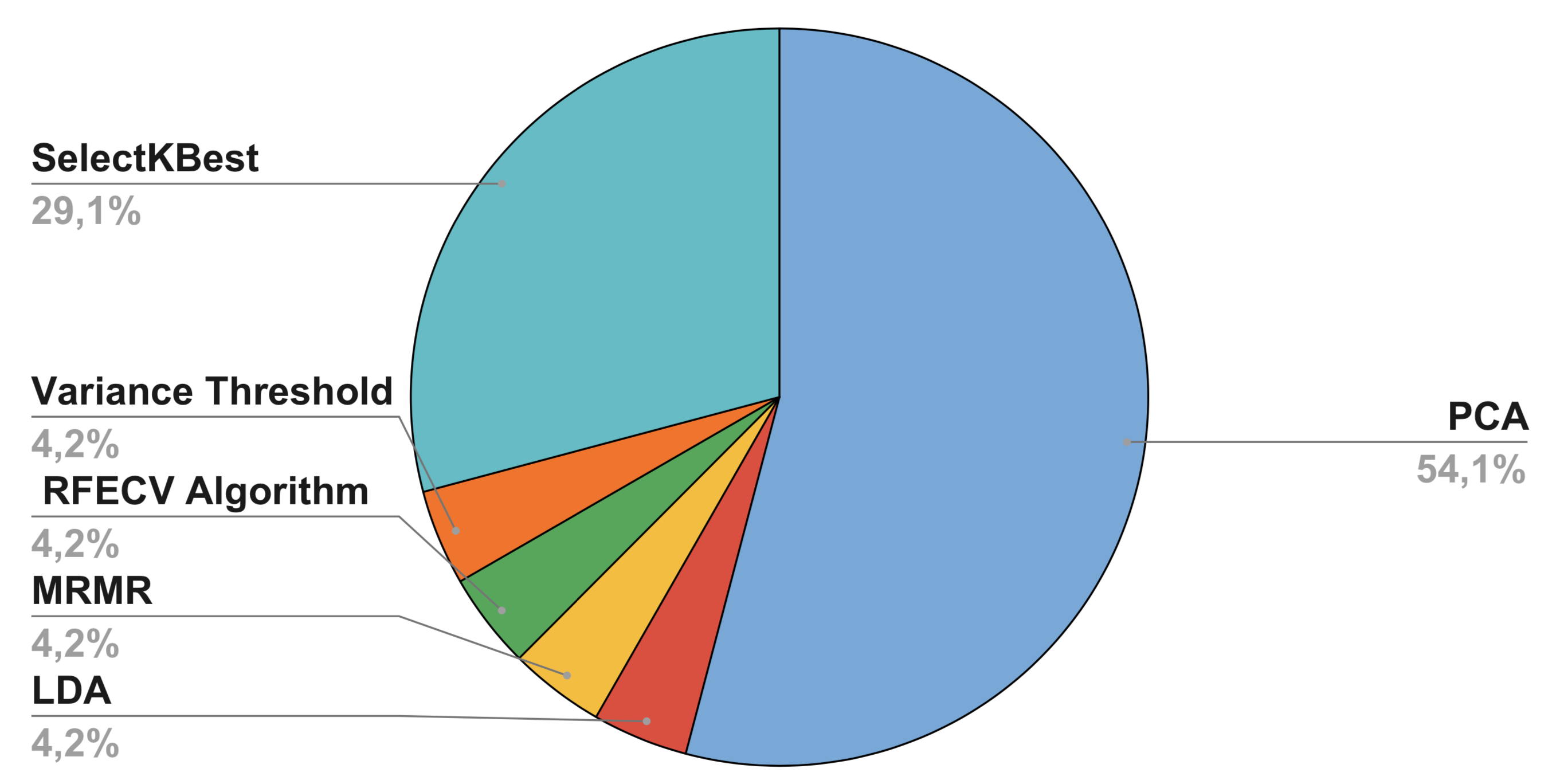}
  \caption{Dimensionality reduction methods used in the brain network classification Kaggle competition for distinguishing between autistic and healthy subjects.}
  \label{fig:reduction}
\end{figure}

According to Figure \ref{fig:reduction}, more than half of the competitors have opted for Principal Component Analysis (PCA). SelectKBest has been identified as the second most used dimensionality reduction technique, accounting for 29.1\% of total usage. On the other hand, each of other four dimensionality reduction methods (RFECV, Variance Threshold, MRMR, and LDA) has been used by only one team, with a 4.2\% usage rate. 


\textbf{Feature Extraction Techniques} \,\, Principal Component Analysis (PCA) and Linear Discriminant Analysis (LDA) are typical instances of feature extraction methods in machine learning, which synthesize new, but fewer number of features. PCA generates those features in a new feature space by projecting the data samples onto pre-defined axes on which the samples are dispersed most (i.e., maximizing the variance of the projected data in the new space). To find out those axes, each of which is defined by a unique unit vector called a principal component, it utilizes singular value decomposition (SVD) of a data-driven matrix \citep{géron2017hands-on}. Likewise, LDA aims to discover new feature axes on which the samples are most scattered, but additionally it also tries to group the data samples from the same class together, which allows sample separation with respect to different classes \citep{raschka2014linear}. Hence, it can be regarded as a \emph{supervised} version of PCA.

\textbf{Feature Selection Techniques} \,\, SelectKBest, Variance Threshold, Minimum Redundancy Maximum Relevance (MRMR), and Recursive Feature Elimination with Cross Validation (RFECV) are categorized as feature selection methods. Their general aim is to choose the most discriminative subset of original features; however, each of them has a different criterion to figure out the importance rate of the features. To generally overview them, variance threshold technique computes the variance of all features across the whole dataset, and eliminates those whose variance value is less than a pre-determined threshold \citep{scikit-learn}. Likewise, SelectKBest is a similar approach, where the top $k$ features with the highest scores are selected by a scoring function (e.g., chi-square or f-score) \citep{scikit-learn}. On the other hand, MRMR algorithm takes the basis of correlation between features and labels of data samples for space reduction rather than scoring those features depending on a particular metric or a score generator. At first, it eliminates the features less correlated with the ground truth labels, since they are poorly effective in disentangling classes, which defines the maximum relevance step. Next, it discards one member of any feature pair highly-correlated with each other by following redundancy principal that if two features are highly correlated, one of them can substitute the other one. While carrying out this two-step process, MRMR utilizes a correlation matrix, exhibiting mutual dependency of features and labels \citep{article}. RFECV method differs from those three in the sense that it needs to employ a learning model due to its wrapper based characteristics. It shrinks the original feature set in each recursive operation in which the base learner is trained with given training dataset, and $k$ number of features with the lowest discriminative scores are discarded. Cross-validation enables this algorithm to determine the most effective dimension of feature space, that is the number of features that will be kept to train the classification model \citep{guyon2002gene}.

\subsubsection{Learning Models}


Learning models are pre-experienced systems in machine learning domain that can classify problem-specific items or estimate target outputs of those items depending on the type of the problem. To obtain this ability, they try to capture and figure out the hidden relationship in the dataset, which is provided for the learning procedure of the models as ready experience. In some cases, they may fail to learn from the dataset, and go through overfitting or underfitting. In this competition, almost all teams have utilized dimensionality reduction methods to circumvent the curse of dimensionality, that is one of those cases, and to make this learning process easier for their models. However, the most of the groups have achieved very low classification results in the start of the competition in spite of applying preprocessing and dimensionality reduction techniques. This has automatically led the competitors to use more advanced learning techniques such as the collaboration of multiple predictors or learners, namely ensemble learning techniques.

\begin{figure}[ht!]
\centering
  \includegraphics[width=0.8\textwidth]{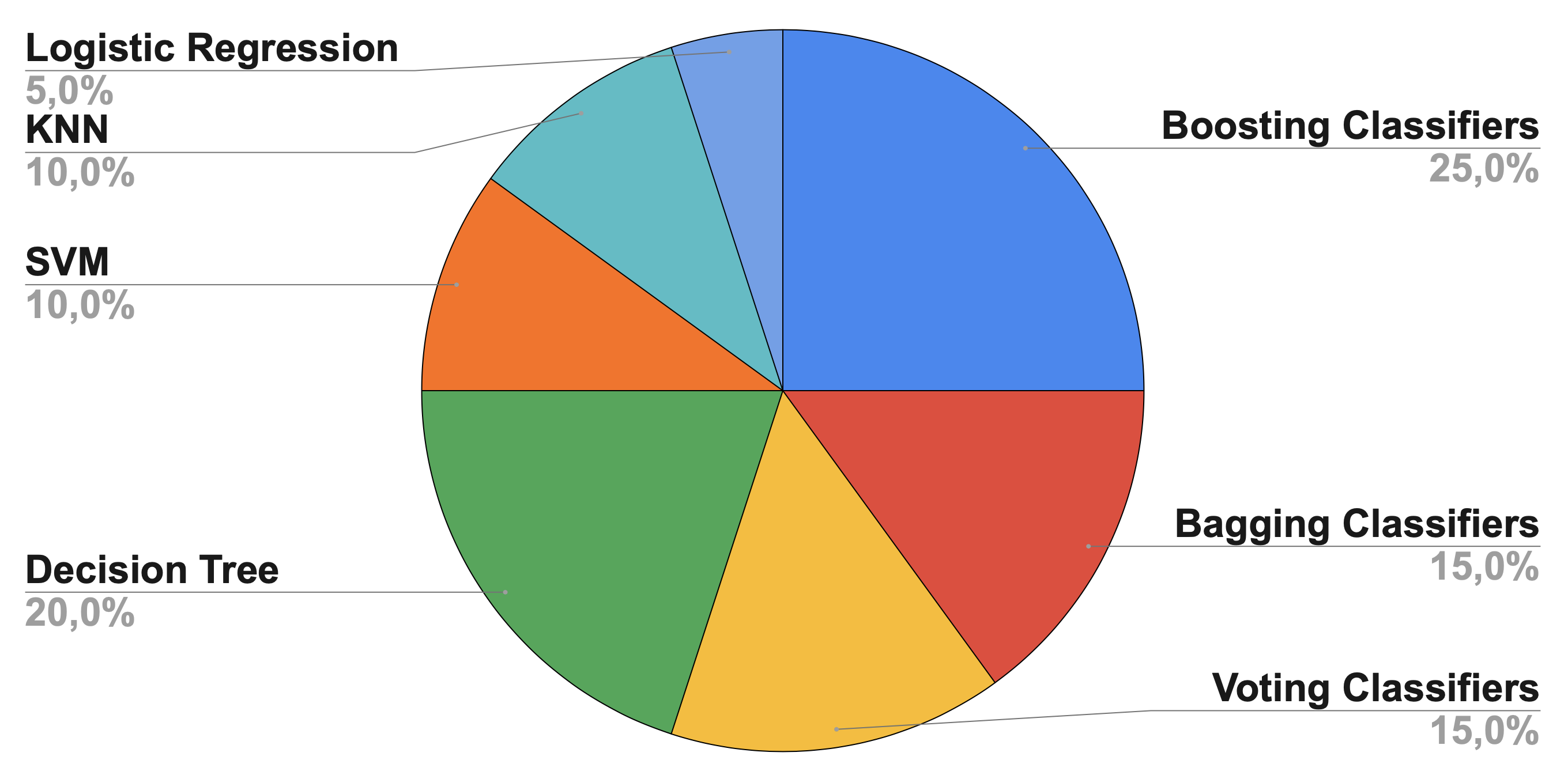}
  \caption{Distribution of the learning methods used in the Kaggle competition for classifying autistic and healthy subjects using cortical morphological networks.}
  \label{fig:learning}
\end{figure}

As it is depicted in Figure \ref{fig:learning}, 55\% of the 20 teams in this competition have preferred to utilize ensemble learning models, which are mainly categorized into Boosting, Bagging and Voting systems. On the other hand, single-learner systems have comprised 45\% of total usage. To examine them in detail, while one-fifth of the competitors have chosen decision tree as their major model, the proportions of $K$-Nearest Neighbor (KNN), and Support Vector Machines (SVM) users have been equivalent (10\% for both). Logistic regression has been the least preferred classification technique for the diagnosis of ASD with solely 5\% of usage.

\textbf{Ensemble Learning} \,\, The fundamental principle that ensemble learning adopts is to confer with the aggregation of learning models rather than trusting in the classification performance that only one of them achieves. However, the essential point in this principle is how each of the learning models in an ensemble system is allowed to learn distinct and unique patterns from the dataset, which is called diversity. 
First approach used to fulfill this diversification is to expose each base learner in an ensemble system to a subset of the training set, not whole distribution in training stage, which is a typical data sampling technique for each one of them. In this strategy, since the learning process is diversified among base learners by different local subsets of the training data, same type of base learner is used to set up the underlying structure of ensemble system. On the other hand, to promote diversity in learning, it is also possible that different kinds of base learners are trained with the entire training set. The main categories of ensemble learning (boosting, bagging, and voting) are based on both approaches \citep{zhang2012ensemble}.

Bagging systems accommodate multiple homogeneous base learners to cooperate with one another. In a typical bagging system, a subset of the training data is constructed for training each base learner using uniform sampling. Both operations, construction of those subsets and base learners' being trained with them, are implemented in parallel thanks to the non-existence of direct dependency between subsets or base learners. In general, bagging systems provide a `replacement' option, which enables a data sample to be added to more than one training subset. On the contrary, pasting algorithm \citep{breiman1999pasting} restricts each data sample to belong to only one training subset. In the testing phase of a bagging system, the majority or average of all predictions made by base learners for testing samples is computed depending on the target problem being classification or regression task respectively \citep{breiman1996bagging}. Two teams in the competition have deployed this ensemble system with the base learners K-Nearest Neighbor and Support Vector Machines, while one team has opted for a specialized version of bagging system for decision trees, called as Random Forest \citep{breiman2001random}.

Boosting algorithms combine multiple homogeneous base learners to work in collaboration just like bagging systems; however, they define relations and dependencies between those base learners to gradually improve the ensemble performance. In particular, adaptive boosting, a typical instance of boosting methods, trains each base learner with a particular subset of the training data; however, the probability of the samples' being chosen for that subset is correlated with their  weight values, which are assigned to them equally at the beginning of the training stage. When a base learner is trained with its corresponding subset, its generalization is tested on the entire training set, and the weights of misclassified samples are increased to make sure that those will be included in the training subset of the next base learner. In this way, each base learner is guaranteed to be trained with mistakes that have been made by previous learners, thereby increasing the difficulty of the classification task at hand for the next learner which enhances its robustness. The main purpose is to expose the learners to the weaknesses of the overall system so that they can boost themselves and become stronger against those weaknesses. On the other hand, in the testing phase, a weighted majority voting over the predictions that are made by the base learners is adopted to estimate the final decision of the boosting system in labeling the testing samples \citep{freund1995desicion}. Apart from AdaBoost, Gradient Boosting \citep{friedman2001greedy} and XGBoost (Extreme Gradient Boosting) \citep{chen2016xgboost} are also variants of boosting algorithms that have been utilized by the competitors. They provide more generalized schemes to boosting in which additive improvement and gradient descent are combined. 


These two ensemble learning techniques (bagging and boosting) are alike in that both of them carry out diversification over subsets of training samples, while employing a single type of base learner. On the contrary, voting techniques, a third type of ensemble learning, fuse different types of learners over the entire training set. The controversial issue that is encountered in voting ensembles is how the predictions of different sorts of learners are combined during the testing stage. In fact, although a variety of approaches for the combination of base classifiers is presented by \citep{kittler1998combining}, the most preferred one in this competition is majority voting, also called as hard-voting. In majority voting principle, all base learners in the voting ensemble make their predictions about which class a testing sample belongs to, and the final vote (i.e., predicted label) will be the most frequent or `agreed upon' label outputted by all learners \citep{zhang2012ensemble}.

\section{Results}

\subsection{Evaluation Dataset} 

\textbf{Cortical morphological network (CMN) construction} \,\, We used Autism Brain Imaging Data Exchange (ABIDE I) dataset\footnote{\url{http://preprocessed-connectomes-project.org/abide/}}, comprising 200 subjects (100 normal controls (NC) and 100 ASD). \citep{Di-Martino:2014aa}. Each subject has a structural T1-w MRI. We used FreeSurfer \citep{Fischl:2012aa} to extract both right and left hemispheres (RH and LH) for each subject, then parcellate each into $N_r = 35$ cortical regions of interest using Desikan-Killiany Atlas. For each subject, we generate a CMN using the maximum principal curvature (MPC) as a cortical attribute as introduced in \citep{mahjoub2018brain}. Specifically, we first compute the average MPC in each cortical ROI. To define the morphological connection between two ROIs, we compute the absolute distance between average MPC in both ROIs. In a CMN, when two ROIs $R_i$ and $R_j$ become more similar in morphology, their morphological connectivity nears zero. By vectorizing the off-diagonal upper triangular part of each CMN, we generate a connectivity vector of size $N_c = N_r \times (N_r-1)/2$ (i.e., 595 connectivities for $N_r=35$).

\textbf{Random partition of the evaluation dataset} \,\, Training and testing sets which have been provided for competing teams were composed of 120 and 80 samples respectively, each of which refers to a cortical morphological network. The testing set has been equally divided into public and private parts in which the competitors were allowed to check the generalizability of their learning models over only public part. In other words, it was designed that they could monitor accuracy results of their models for solely the public part of the whole unseen testing set during the competition, while the accuracy score of predictions over the private subset was kept  inaccessible for even testing until the end of the competition. Upon the end of the competition, the default ML frameworks proposed by the teams were tested over the entire testing set. Adopting such an evaluation strategy primarily aimed to first avoid the overfitting stemming from the teams' behavior of exhaustive submissions to Kaggle to monitor the accuracy results of their ML models, and second evaluate the generalizability of the designed models on an extra testing dataset they did not have access to for model training and testing.

\subsection{Performance Metrics} 

We evaluated the performance of ML pipelines for ASD/NC classification by using the results derived from the public and private testing subsets in terms of accuracy, sensitivity, and specificity metrics. The public test set was partially accessible to the competitors during the competition so that they could check their model accuracy, whereas the private one was completely kept hidden to assess the generalizability of the models after the competition. By using these two test sets, public and private accuracy, sensitivity, and specificity values were computed for the method of each team, and then those two results were averaged for all three metrics. All participating teams were ranked in three different ways, each of which takes the basis of one type of average measurement, and the sum of three ranks determined the final rank of each team in the competition. Table~\ref{tab:results_table} displays the classification results for the top 20 competitors and Table~\ref{tab:methods_details} details the components of the 20 designed ML pipelines.








\begin{table}[h!]
\centering
\resizebox{1.0\textwidth}{!}{%
\begin{tabular}{@{}lllllllll@{}}
\toprule
\textbf{Team ID} & \textbf{Accuracy} & \textbf{Sensitivity} & \textbf{Specificity} & \textbf{\begin{tabular}[c]{@{}l@{}}Accuracy\\ Rank\end{tabular}} & \textbf{\begin{tabular}[c]{@{}l@{}}Sensitivity\\ Rank\end{tabular}} & \textbf{\begin{tabular}[c]{@{}l@{}}Specificity\\ Rank\end{tabular}} & \textbf{\begin{tabular}[c]{@{}l@{}}Sum of \\ Ranks\end{tabular}} & \textbf{\begin{tabular}[c]{@{}l@{}}Final\\ Rank\end{tabular}} \\ \midrule
1                & 0.700             & 0.725                & 0.675                & 1                                                                & 2                                                                   & 5                                                                   & 8                                                                & 1                                                             \\
2                & 0.638             & 0.625                & 0.650                & 2                                                                & 5                                                                   & 6                                                                   & 13                                                               & 2                                                             \\
3                & 0.600             & 0.425                & 0.775                & 3                                                                & 15                                                                  & 2                                                                   & 20                                                               & 3                                                             \\
10               & 0.600             & 0.700                & 0.500                & 3                                                                & 3                                                                   & 15                                                                  & 21                                                               & 4                                                             \\
5                & 0.588             & 0.650                & 0.525                & 5                                                                & 4                                                                   & 13                                                                  & 22                                                               & 5                                                             \\
8                & 0.563             & 0.525                & 0.600                & 6                                                                & 9                                                                   & 7                                                                   & 22                                                               & 5                                                             \\
7                & 0.563             & 0.625                & 0.500                & 6                                                                & 5                                                                   & 15                                                                  & 26                                                               & 7                                                             \\
11               & 0.550             & 0.500                & 0.600                & 8                                                                & 10                                                                  & 8                                                                   & 26                                                               & 7                                                             \\
13               & 0.538             & 0.500                & 0.575                & 10                                                               & 10                                                                  & 9                                                                   & 29                                                               & 9                                                             \\
9                & 0.550             & 0.350                & 0.750                & 8                                                                & 19                                                                  & 3                                                                   & 30                                                               & 10                                                            \\
12               & 0.538             & 0.375                & 0.700                & 10                                                               & 17                                                                  & 4                                                                   & 31                                                               & 11                                                            \\
4                & 0.538             & 0.925                & 0.150                & 10                                                               & 1                                                                   & 20                                                                  & 31                                                               & 11                                                            \\
6                & 0.525             & 0.050                & 1.000                & 13                                                               & 20                                                                  & 1                                                                   & 34                                                               & 13                                                            \\
14               & 0.525             & 0.550                & 0.500                & 13                                                               & 8                                                                   & 15                                                                  & 36                                                               & 14                                                            \\
17               & 0.525             & 0.575                & 0.475                & 13                                                               & 7                                                                   & 19                                                                  & 39                                                               & 15                                                            \\
16               & 0.513             & 0.450                & 0.575                & 16                                                               & 14                                                                  & 9                                                                   & 39                                                               & 15                                                            \\
18               & 0.500             & 0.425                & 0.575                & 17                                                               & 15                                                                  & 9                                                                   & 41                                                               & 17                                                            \\
19               & 0.500             & 0.475                & 0.525                & 17                                                               & 12                                                                  & 13                                                                  & 42                                                               & 18                                                            \\
15               & 0.475             & 0.375                & 0.575                & 20                                                               & 17                                                                  & 9                                                                   & 46                                                               & 19                                                            \\
20               & 0.488             & 0.475                & 0.500                & 19                                                               & 13                                                                  & 15                                                                  & 47                                                               & 20                                                            \\ \bottomrule
\end{tabular}%
}
\vspace{3pt}
\caption{The ranks of the top 20 competitors.}
\label{tab:results_table}
\end{table}



\subsection{Best performing methodologies} 

\begin{figure}[ht!]
\centering
  \fbox{\includegraphics[width=\textwidth]{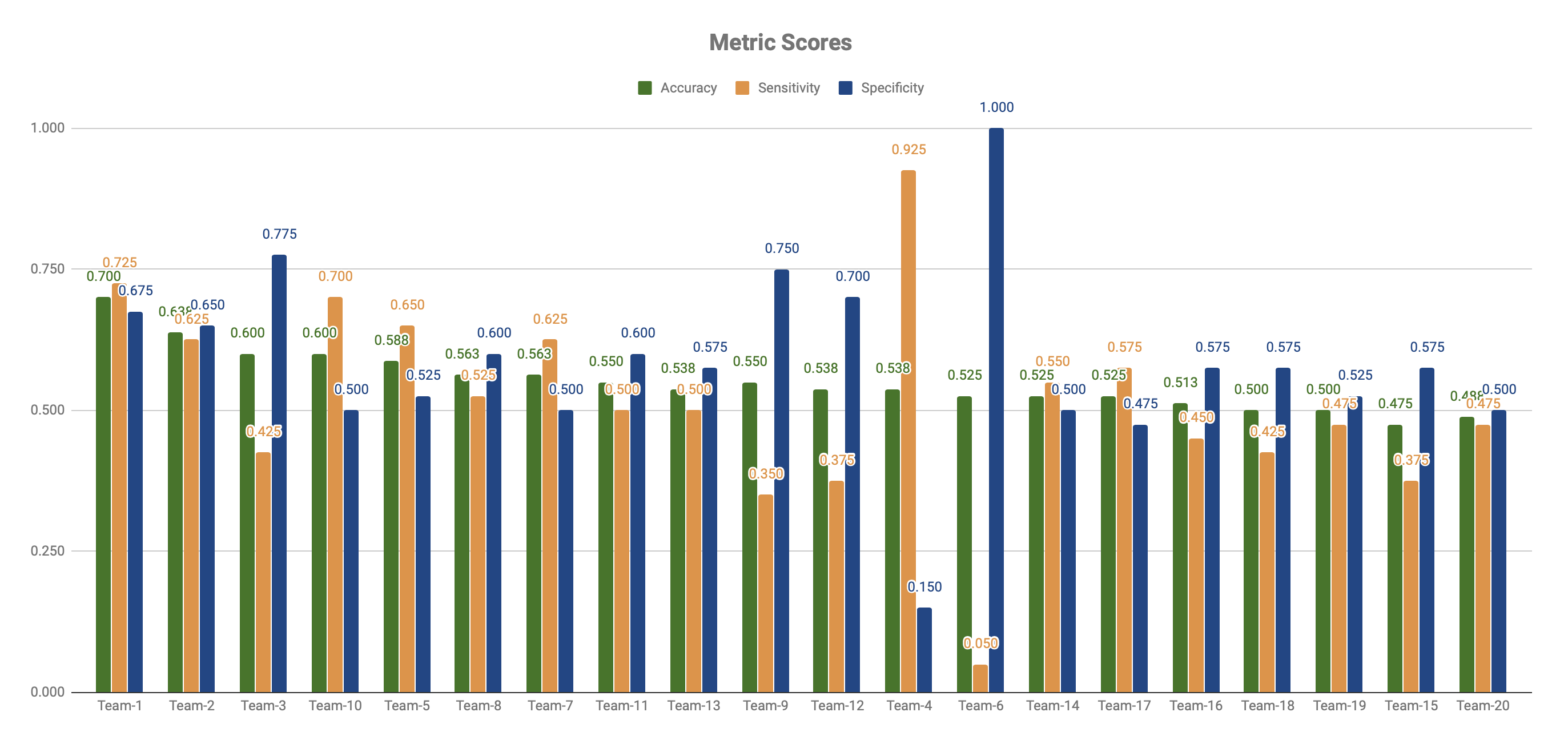}}
\caption{Evaluation scores of the 20 participating teams.}
\label{fig:avgmeas}
\end{figure}

\begin{figure}[ht!]
\centering
  \fbox{\includegraphics[width=\textwidth]{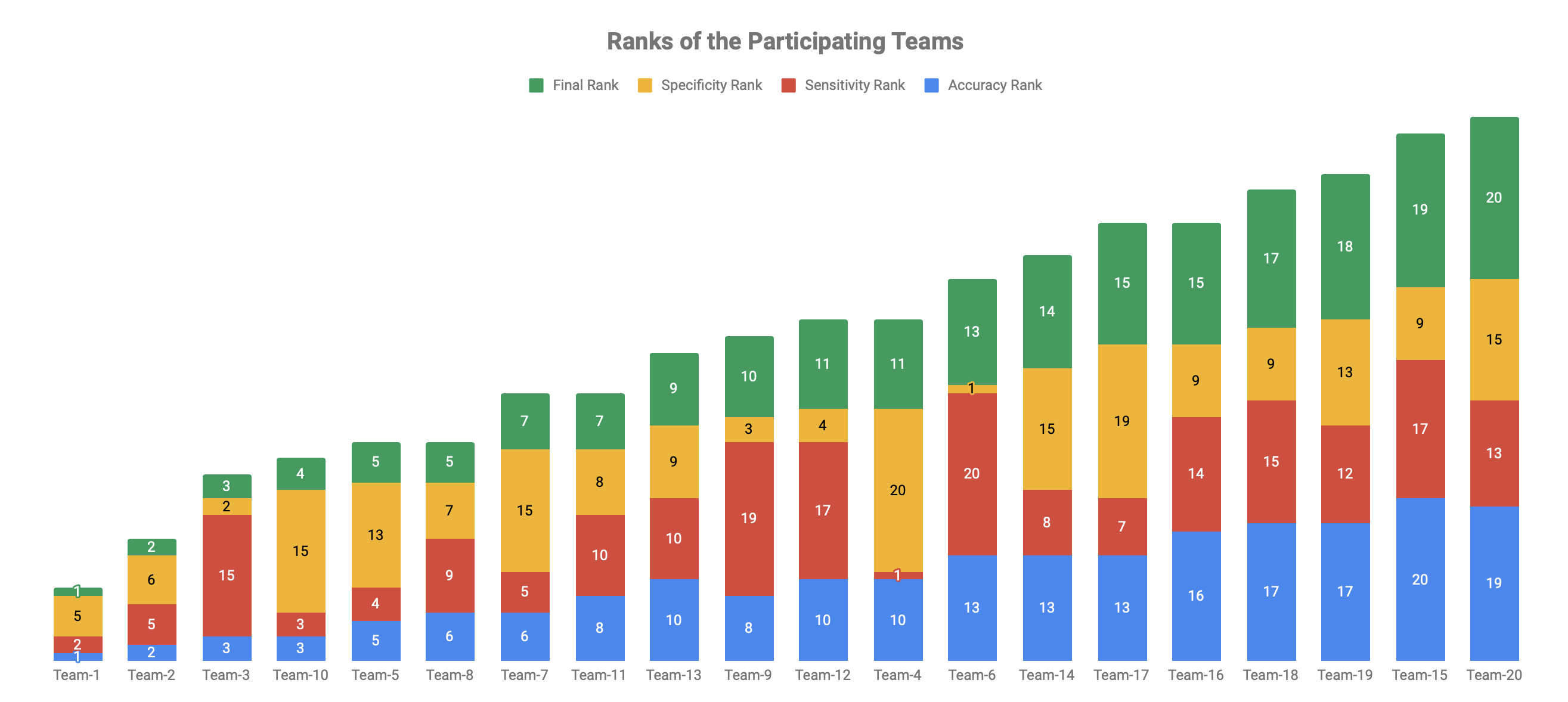}}
\caption{The ranks of the 20 participating teams.}
\label{fig:ranks}
\end{figure}

Evaluation scores of ML pipelines designed for ASD/NC classification are illustrated in Figure \ref{fig:avgmeas}. These performance measurements comprise accuracy, sensitivity, and specificity metrics. On the other hand, Figure \ref{fig:ranks} gives information about how those pipelines are ranked with respect to those three evaluation scores. In both charts, the teams are ordered from left to right depending on ascending overall competition rank. As it is depicted by Figure \ref{fig:avgmeas}, while the accuracy scores of the top three teams varied between 0.6 and 0.7, the opposite group (the worse ones) was stabilized in approximately 0.5 accuracy rate.

In this competition, two major methodologies were adopted in the design of the competing learning models, which are single-learner systems and ensemble learning. Support Vector Machines (SVM) \citep{chang2011libsvm}, Logistic Regression \citep{hosmer2013applied}, K-Nearest Neighbor (KNN) \citep{Peterson:2009}, and Decision Tree \citep{breiman1984classification} are single-learner based methods used in the competition. The results revealed by the charts in Figures \ref{fig:avgmeas} and \ref{fig:ranks} highlight which one of these two methodologies were more prominent and achieved remarkable results. Notably, the two best performing teams with classification accuracies of 0.70 and 0.638, respectively, and a quite balanced sensitivity-specificity metric scores, opted for Gradient Boosting technique in ensemble learning. Likewise, team 3 with a 0.6 classification accuracy used one of the ensemble learning methods, that is a voting system of multiple predictors with a majority voting option. This reveals that building a synergy between different learners has a higher fault tolerance, and consequently, it surpasses the weaknesses of single learner systems. However, this does not entail that using an ensemble learning method constantly guarantees the best results or a better generalizability than single-learners. For instance, team 10 has chosen decision tree classifier as default model, and left some ensemble users behind by getting ranked fourth in the competition. In addition to this, although team 15 has benefited from bagging algorithm in ensemble learning with SVM base classifier, it was in the penultimate position with solely 0.475 accuracy rate. 

\renewcommand{\arraystretch}{1.8}
\begin{sidewaystable}
  \centering
  \resizebox{1.0\textwidth}{!}{%
\begin{tabular}{|l|>{\columncolor[HTML]{f4e9dc}}c|>{\columncolor[HTML]{f4e9dc}}c|>{\columncolor[HTML]{f4e9dc}}c|>{\columncolor[HTML]{f4e9dc}}c|>{\columncolor[HTML]{d8e3e3}}c|>{\columncolor[HTML]{d8e3e3}}c|>{\columncolor[HTML]{d8e3e3}}c|>{\columncolor[HTML]{d8e3e3}}c|>{\columncolor[HTML]{d8e3e3}}c|>{\columncolor[HTML]{d8e3e3}}c|>{\columncolor[HTML]{d8e3e3}}c|>{\columncolor[HTML]{d8e3e3}}c|>{\columncolor[HTML]{f7edec}}c|>{\columncolor[HTML]{f7edec}}c|>{\columncolor[HTML]{f7edec}}c|>{\columncolor[HTML]{f7edec}}c|>{\columncolor[HTML]{f7edec}}c|>{\columncolor[HTML]{f7edec}}c|>{\columncolor[HTML]{f7edec}}c|>{\columncolor[HTML]{f7edec}}c|>{\columncolor[HTML]{f7edec}}c|>{\columncolor[HTML]{f7edec}}c|>{\columncolor[HTML]{f7edec}}c|>{\columncolor[HTML]{f7edec}}c|>{\columncolor[HTML]{f7edec}}c|>{\columncolor[HTML]{f7edec}}c|>{\columncolor[HTML]{f7edec}}c|}
\hline
\textbf{Teams} & \textbf{MMS} & \textbf{SS} & \textbf{CFE} & \textbf{IF} & \textbf{CBE} & \textbf{EHCF} & \textbf{SKBA} & \textbf{LDA} & \textbf{PCA} & \textbf{MRMR} & \textbf{VT} & \textbf{RFECV} & \textbf{VC} & \textbf{BC} & \textbf{RF} & \textbf{AB} & \textbf{GB} & \textbf{XGB} & \textbf{SGD} & \textbf{LR} & \textbf{SVC} & \textbf{DTC} & \textbf{DTR} & \textbf{KNN} & \textbf{LDA} & \textbf{QDA} & \textbf{NB} \\ \hline
Team-1 & & & \cellcolor[HTML]{F4BF81} & & & & \cellcolor[HTML]{68CBD0} & & & & & & & & & & \cellcolor[HTML]{F49C92} & & & & & & & & & & \\ \hline
Team-2 & & & & & & & & & & \cellcolor[HTML]{68CBD0} & & & & & & & \cellcolor[HTML]{F49C92} & & & & & & & & & & \\ \hline
Team-3 & & \cellcolor[HTML]{F4BF81} & & & & & & & \cellcolor[HTML]{68CBD0} & & & & \cellcolor[HTML]{F49C92} & & & \cellcolor[HTML]{F2C3BE} VC & \cellcolor[HTML]{F2C3BE} VC & \cellcolor[HTML]{F2C3BE} VC & & & \cellcolor[HTML]{F2C3BE} AB & & & & & & \\ \hline
Team-10 & & & \cellcolor[HTML]{F4BF81} & & & & & & \cellcolor[HTML]{68CBD0} & & & & & & & & & & & & & \cellcolor[HTML]{F49C92} & & & & & \\ \hline
Team-5 & \cellcolor[HTML]{F4BF81} & & & & & \cellcolor[HTML]{68CBD0} & \cellcolor[HTML]{68CBD0} & & & & & & & & & & & \cellcolor[HTML]{F49C92} & & & & & & & & & \\ \hline
Team-8 & & \cellcolor[HTML]{F4BF81} & & & & & & & \cellcolor[HTML]{68CBD0} & & & & & & & & & & & \cellcolor[HTML]{F49C92} & & & & & & & \\ \hline
Team-7 & & & \cellcolor[HTML]{F4BF81} & & & & \cellcolor[HTML]{68CBD0} & & \cellcolor[HTML]{68CBD0} & & & & & & & & & & & & & \cellcolor[HTML]{F49C92} & & & & & \\ \hline
Team-11 & & & & & & & \cellcolor[HTML]{68CBD0} & & & & & & & & & \cellcolor[HTML]{F49C92} & & & & & & \cellcolor[HTML]{F2C3BE} AB & & & & & \\ \hline
Team-13 & & & & & \cellcolor[HTML]{68CBD0} & & \cellcolor[HTML]{68CBD0} & & & & & & & & & \cellcolor[HTML]{F49C92} & & & & & & \cellcolor[HTML]{F2C3BE} AB & & & & & \\ \hline
Team-9 & \cellcolor[HTML]{F4BF81} & & & & & & & & \cellcolor[HTML]{68CBD0} & & & & & \cellcolor[HTML]{F49C92} & & & & & & & & & & \cellcolor[HTML]{F2C3BE} BC & & & \\ \hline
Team-4 & &  \cellcolor[HTML]{F4BF81} & & & & & & & \cellcolor[HTML]{68CBD0} & & & & & & & & & & & & & \cellcolor[HTML]{F49C92} & & & & & \\ \hline
Team-12 & & & & & & & & & \cellcolor[HTML]{68CBD0} & & & & & & & & & & & & \cellcolor[HTML]{F49C92} & & & & & & \\ \hline
Team-6 & & \cellcolor[HTML]{F4BF81} & & & & & & & \cellcolor[HTML]{68CBD0} & & & & & & & & & & & & \cellcolor[HTML]{F49C92} & & & & & & \\ \hline
Team-14 & & & & & & & & & \cellcolor[HTML]{68CBD0} & & & & & & \cellcolor[HTML]{F49C92} & & & & & & & & & & & & \\ \hline
Team-16 & & & & & & & \cellcolor[HTML]{68CBD0} & & \cellcolor[HTML]{68CBD0} & & & & & & & & & & & & & & & \cellcolor[HTML]{F49C92} & & & \\ \hline
Team-17 & & & & & & & & \cellcolor[HTML]{68CBD0} & & & & & \cellcolor[HTML]{F49C92} & & \cellcolor[HTML]{F2C3BE} VC & & \cellcolor[HTML]{F2C3BE} VC & & \cellcolor[HTML]{F2C3BE} VC & & \cellcolor[HTML]{F2C3BE} VC & & & & & & \cellcolor[HTML]{F2C3BE} VC \\ \hline
Team-18 & & & & & & & \cellcolor[HTML]{68CBD0} & & \cellcolor[HTML]{68CBD0} & & \cellcolor[HTML]{68CBD0} & & \cellcolor[HTML]{F49C92} & & & & & & & & \cellcolor[HTML]{F2C3BE} VC & \cellcolor[HTML]{F2C3BE} VC & & \cellcolor[HTML]{F2C3BE} VC & \cellcolor[HTML]{F2C3BE} VC & \cellcolor[HTML]{F2C3BE} VC & \\ \hline
Team-19 & & & & & & & & & \cellcolor[HTML]{68CBD0} & & & & & & & & & & & & & & & \cellcolor[HTML]{F49C92} & & & \\ \hline
Team-15 & & & & \cellcolor[HTML]{F4BF81} & & & & & & & & \cellcolor[HTML]{68CBD0} & & \cellcolor[HTML]{F49C92} & & & & & & & \cellcolor[HTML]{F2C3BE} BC & & & & & & \\ \hline
Team-20 & & & \cellcolor[HTML]{F4BF81} & & & & & & \cellcolor[HTML]{68CBD0} & & & & & & & & & & & & & & \cellcolor[HTML]{F49C92} & & & & \\ \hline
\end{tabular}
}
\vspace{6pt}
\caption{\scriptsize The preprocessing (orange), dimensionality reduction (blue) and learning methods (red) used by the top 20 teams. The teams were sorted based on the final ranking. Lighter-colored boxes in learning methods indicate the sub-methods used in conjunction with the ensemble methods written inside them. The method names are abbreviated as: Min-Max Scaling (MMS), Standard Scaling (SS), Constant, Features Elimination (CFE), Isolation Forest (IF), Correlation Based Elimination (CBE), Elimination of Highly-Correlated Features (EHCF), SelectKBest Algorithm (SKBA), Linear Discriminant Analysis (LDA), Principal Component Analysis (PCA), Minimum Redundancy Maximum Relevance (MRMR), Variance Threshold (VT), Recursive Feature Elimination and Cross-validated Selection (RFECV), Voting Classifier (VC), Bagging Classifier (BC), Random Forest (RF), AdaBoost (AB), Gradient, Boosting (GB), XGBoost (XGB), Stochastic, Gradient Descent (SGD), Logistic Regression (LR), C-Support Vector Classification (SVC), Decision Tree Classifier (DTC), Decision Tree Regressor (DTR), K-Nearest Neighbors (KNN), Linear Discriminant Analysis (LDA), Quadratic Discriminant Analysis (QDA), Naive Bayes (NB).}
\label{tab:methods_details}%
\end{sidewaystable}

\section{Discussion}




\textbf{Overview} \,\, Distinguishing the autistic from the typical brain using structural T1-weighted MRI is still challenging due to the disorder heterogeneity as well as subtle changes in the brain structure. Cortical morphological networks derived merely from structural T1-w MRI were used in this Kaggle competition to design robust and accurate ML frameworks for ASD diagnosis.

Crowdsourcing by means of community competitions has been widely used as a framework to address complex scientific problems in various areas such as neuroimaging \citep{bron2015standardized, rodriguez2016} and genomics \citep{belcastro2018, marbach2012wisdom}. To fill the gap in  ASD disagnosis using morphological cortical morphological networks, we organized an in-class competition via Kaggle to investigate a wide spectrum of ML pipelines and scrutinize their strengths and limitations. We encouraged the participants to develop their genuine ML model pipelines collaboratively by enabling to join teams of at most 3 members. While the competition was ongoing, the teams were able to only access the classification accuracy on the public part of the test set. In this way, they were pushed to generate models that better generalize to unseen subjects. At the end of the competition, the best 20 submissions in terms of public testing accuracy were included in this study for further examination. The final ranking was determined based on the sum of the ranks for each evaluation metric using both public and private testing sets. More importantly, we provide the open source codes in Python of the top 20 competing teams at \url{https://github.com/basiralab/BrainNet-ML-ToolBox}. Although, these sources codes were primarily designed for ASD diagnosis using CMN datasets, they can be utilized for brain network classification in general. 

\textbf{Best and worst performing ML algorithms} \,\,This competition enabled the comparison of different machine learning methodologies for distinguishing between autistic and healthy subjects using cortical morphological networks. Table \ref{tab:methods_details} shows the preprocessing and classification methods used by each individual team. The best performing team (team-1) achieved an overall accuracy, sensitivity, and specificity scores of 0.7, 0.725, and 0.675 respectively (Table \ref{tab:results_table}) using univariate feature selection method (selectKBest) based on univariate chi-square statistical tests, followed by Gradient Boosting classification method. The second team (team-2) applied MRMR algorithm for exploring the better representative features, then used Gradient Boosting model for classifying subjects, and achieved 0.638, 0.625, and 0.650 scores in accuracy, sensitivity, and specificity metrics respectively. While the first two teams have opted for the Gradient Boosting technique, the team in the third-order has benefited from voting of multiple predictors with the majority voting option. Team-3 and team-10 took 3rd and 4th places respectively in the final ranking with the same accuracy of 0.6. Team-3 preferred an ensemble voting scheme involving different boosting algorithms like AdaBoost and XGBoost after shrunk the feature space using PCA. In this way, they obtained high specificity (0.775) but quite low sensitivity (0.425), which means successful identification of the normal controls but overlooking actual autism cases significantly. Unlike the top three teams, Team-10 preferred to use merely decision tree classifier instead of ensemble methods, and achieved enhanced sensitivity (0.7) but poor sensitivity (0.5) in contrary to Team-3. Other teams did not get remarkable results since their accuracy remained below 0.6 and the average of sensitivity-specificity was low.

Although both are expected to be high, there is typically an inherent trade‐off between sensitivity and specificity in ASD diagnosis due to its complexity \citep{randall2018}. A prediction with high sensitivity but low specificity cause overdiagnosis. On the other hand, low sensitivity but high specificity could result in overlooking actual autism subjects. High sensitivity can be favorable in population-based screening (level one) since it is desired to identify the maximum number of subjects at risk. Besides, specificity is more important in further evaluation (level two) to be applied to subjects that are categorized as at-risk previously \citep{lang2016early}. Because high specificity enables a model to discriminate the subjects not affected by ASD more accurately. Therefore, applying a classification method with high specificity followed by another with high specificity value can enhance the classification performance. 

Bagging and boosting algorithms as ensemble methods have been successfully applied in high dimensional data classification problems \citep{pappu2014high}. Especially, boosting methods such as GradientBoosting and XGBoost showed their success in public challenges among many other methods \citep{chen2016xgboost}. In this challenge, the first two best performing teams (team-1 and team-2) exploited GB and outperformed other methodologies. Team-3 at 3rd rank used voting classifier in conjunction with different boosting methods such as AB, GB and XGB. The results clearly show that the ensemble methods can tackle the high dimensionality problem of CMNs better than individual traditional methods. The success of the ensemble methodology is rooted in the principle of combining weak classifiers as to complement each other's weaknesses. The final combination model hopefully yields improved generalizability and robustness over a single model.

However, team-10 was the only team in the top 5 teams, which preferred DT as a single classifier (Table \ref{tab:methods_details}). Mostly, conventional classification models using high-dimensional data are more likely to overfit data than the ensemble methods \citep{pappu2014high}. Hence, for single models, the use of dimensionality reduction techniques such as feature selection or feature extraction are of greater importance than for ensemble methods. Team-10 applied PCA method after eliminating  constant features and used 60 extracted new features to train their model (Table \ref{tab:methods_details}). In this way, they outperformed some ensemble methods including XGBoost and Adaboost. Team-4 also applied DT with PCA but obtained poorer results (11th) unlike team-10 (Figure \ref{fig:ranks}). The main difference in implementations is the number of principal components used. Team-4 used 16 principle components (i.e., new extracted features) while team-10 used 60. This indicates how the optimization of the dimensionality reduction method affects the results.

On the other hand, the ML pipelines in the lower half of ranking mostly suffered from the absence of preprocessing techniques (Table \ref{tab:methods_details}). Team-14, team-16, team-17, and team-18 did not use any preprocessing techniques and the PCA/LDA dimensionality reduction methods they used can heavily be affected by redundant constant features which were not removed. Similarly, 8 of the last 10 teams used PCA but only 2 teams among them have scaled the dataset prior to the application of PCA, which can worsen the model performance significantly.

Although the ensemble methods achieved remarkable results in this challenge, team-14 which used RF, team-17 and team-18 which used VC with various classifier models, and team-15 which used BC achieved poor results (Figure \ref{fig:ranks}). In addition to the aforementioned lack of data preprocessing, team-18 exploited 3 different dimensionality reduction methods together and this obviously negatively affected the prediction accuracy. Team-14 employed RF model with PCA but they preposterously excluded 20 training samples. One of the issues of the ML framework designed by team-15 was probably related to shrinking sample size using IF method. They excluded 7 samples detected as an outlier by IF and trained the model with the remaining 113 samples. Since the size of the dataset is not large, this might have a significant effect on the model performance. Furthermore, they used RFECV based on SVM, which is an iterative method to find the best features, then exploited bagging with SVC to classify subjects without normalizing the features. Making such ML decisions should be rigorously justified and well thought out in future studies.

Another factor having a substantial effect on the ML model performance is parameter tuning. The top 3 teams  leveraged tuning techniques to enhance the predictive power of their models. Team-1 used Python implementation \citep{simlrpy} of SIMLR framework \citep{wang2017visualization}, which is a novel similarity-learning framework to learn the similarity between data samples for dimensionality reduction, clustering, and visualization. Team-2 estimated the optimal parameters based on empirical tuning. Team-3 applied a grid search technique \citep{bergstra2012random} for tuning model parameters.

The problem of ASD diagnosis based on CMNs is characterized by high dimensional data where the number of connectivity features $N_c$ is far greater than subjects ($ N_c >> N$). High dimensionality is a common problem in learning from data, known as the curse of dimensionality problem. Because of the lack of enough subjects to decipher the relation between features, the models have difficulty in generalizing the data accurately and tend to overfitting, which frequently results in a decrease in classification performance. To overcome this, removing irrelevant and redundant data is requisite using an appropriate dimensionality reduction technique. Additionally, utilizing a narrow set of selected features reduces model complexity, which leads to a better understanding of the learning process. However, the success of the feature reduction strategies highly depends on the type of data \citep{janecek2008}.
Since the provided training dataset has zero-valued columns, most of the teams initially have cleared out those columns as an initial step. The majority of teams preferred to use various feature extraction techniques such as PCA to decrease the number of features (Table \ref{tab:methods_details}). Fewer teams relied on feature selection techniques such as selectKBest (Team-1, Team-5, etc.) to find better representative subsets of features. 




\textbf{Limitations of adopted ML techniques } \,\, In feature selection, the information may be lost to some extent, since some of the features are discarded. On the other hand, feature extraction techniques such as PCA suffer from tractability. To infer the most important features, they project the high-dimensional original features into a low-dimensional space, which is generally an intractable process that hinders biomarker identification.  
We note that the top 2 best-performing teams used feature selection techniques, SKBA and MRMR (Table \ref{tab:methods_details}), which allow feature tractability.

Furthermore,  dimensionality reduction methods such as PCA, LDA, and ICA are insufficient to capture highly-nonlinear relationships among variables. Using nonlinear dimensionality reduction methods \citep{tenenbaum2000, zhang2014, debnath2018} can enhance classification accuracy. 
Besides, the problem of high dimensionality can be addressed using a deep learning framework \citep{kiarashinejad2020deep, du2018classification}. Deep learning performs non-linear transformation of the original data. This way, it encodes the data from a high-dimensional manifold into a low-dimensional manifold containing more discriminative information. However, using deep learning comprises its own challenges inside \citep{du2018classification}. Its main limitations are of requiring a sufficiently large training dataset. Also it might be prone to overfitting more likely than traditional methods. None of the teams have incorporated deep learning methods since the goal of the competition was to primarily focus on typical ML tools, which are easy to train and hence can be easily utilized by clinicians.

One of the main difficulties encountered in building classification models from high-dimensional data having a limited number of samples is the tendency to overfitting. Although using the appropriate dimensionality reduction technique can  help alleviate this issue, choosing the model, adjusting the parameters and determining the evaluation criteria are of great importance. In this challenge, ensembles models exhibited more robust performance than individual models. In ensemble models, it is expected that the advantages of different methods complement each other and shortcomings are resolved when their predictions are integrated. Most of the teams trusted in the collaboration of multiple models instead of a single model. Boosting and voting classifiers have been the most preferred ensemble methods in the challenge. First and second-ranked teams utilized the Gradient Boosting (GB) classifier, after applying filtering-based feature selection. GB aims to learn from the misclassified samples in the previous step for improving the models to be able to classify them correctly. Thus, GB is prone to overfitting if it is not stopped early enough. 

\textbf{Study limitations and recommendations for future work} \,\, A limitation of this challenge arose in the evaluation phase. Kaggle allows only a single test set and a scoring metric to measure the submission performance. Therefore, we randomly determined a portion of the dataset as test set and used accuracy as a scoring metric. We further circumvented the limitation of a single scoring metric by measuring the other two metrics, sensitivity and specificity on final submissions of predictions. A better evaluation scheme can be set up using cross-validation.

This study provided insights on the discriminative potential of cortical morphological networks in diagnosing autistic spectrum disorders using a large pool of machine learning techniques. However, for this Kaggle challenge, evaluation metrics were only restricted to classification accuracy, sensitivity and specificity. For translational medicine, clinicians can explore the designed algorithms using other evaluation metrics that they regard as important for their diagnosis. In this challenge, we have only utilized CMNs derived from maximum principle curvature of the cortical surface for ASD diagnosis. For a future challenge, one can use multi-view CMNs \citep{mahjoub2018brain,soussia2018,raeper2018cooperative,lisowska2019joint}, where each CMN is derived from a particular cortical measurement, to investigate the discriminative potential of other types of CMNs as well as design computer-aided diagnosis ML frameworks that operates on multi-view brain networks. These can also be combined with clinical scores and other clinical data.

In this study, we did not constraint the participants to use particular ML methods. This limits a thorough investigation of the designed ML pipelines as we might miss a few details on how some hyper-parameters were optimized. Future comparative works can focus on the top performing methods and conduct a more detailed analysis of their inner engineering steps. This will allow to identify the most reproducible and reliable diagnostic ML pipeline for a more focused clinical research. For this reseason, we made the source codes of all teams available on GitHub\footnote{\url{https://github.com/basiralab/BrainNet-ML-ToolBox}}. Note that this Kaggle challenge was restricted to a few number of participants (a classroom), in a future work, we can organize a public challenge as part of an international conference or workshop. As such, we should provide a larger dataset to better cover the heterogeneity of autism  \citep{geschwind2007autism,hull2017resting}.



\section{Conclusion}


Designing accurate and reproducible machine learning frameworks for autistic spectrum disorder diagnosis using brain networks remains highly challenging due to its heterogeneity. A vast number of studies conducted in neuroimaging involving non-invasive brain imaging modalities such as functional MRI (fMRI) and diffusion MRI (dMRI) for diagnosis. However, morphological brain connectivity derived merely from structural T1-w MRI was used seldom, despite the novel insights it might provide about ASD and how it affects the brain \emph{connectional morphology}. In this study, we exploited crowdsourcing by organizing an in-class Kaggle challenge to fill the gap in the utilization of cortical morphological networks for the diagnosis of ASD. The participants were encouraged to develop their innovative model pipelines as a result of collaboration of their teams. Our competition enabled benchmarking a large number  of machine learning methods comprising connectomic data preprocessing and classification steps. Although the training data is \emph{small} and \emph{high-dimensional}, the best performing team (team-1) achieved  0.7, 0.725, and 0.675, and the second team (team-2) achieved 0.638, 0.625, and 0.650 accuracy, sensitivity, and specificity scores, respectively. This demonstrates the discriminative potential that CMNs hold in diagnosing autism. In future large-scale challenges, we can leverage multi-modal brain networks including functional, structural, and morphological networks to design new ML frameworks that can handle multimodal datasets and investigate the relationship between those networks and how they are altered by autism. The designed frameworks in this challenge are generic and can be used for neurological disorder diagnosis using uni-modal (i.e., single type) brain networks. We refer interested readers to our GitHub website \url{https://github.com/basiralab/BrainNet-ML-ToolBox}.


\section{Acknowledgements}
This project has been funded by the 2232 International Fellowship for Outstanding Researchers Program of TUBITAK (Project No:118C288) supporting I. Rekik. However, all scientific contributions made in this project are owned and approved solely by the authors.

\newpage
\bibliography{Kbiblio}

\begin{thebibliography}{60}
\expandafter\ifx\csname natexlab\endcsname\relax\def\natexlab#1{#1}\fi
\expandafter\ifx\csname url\endcsname\relax
  \def\url#1{\texttt{#1}}\fi
\expandafter\ifx\csname urlprefix\endcsname\relax\def\urlprefix{URL }\fi
\providecommand{\eprint}[2][]{\url{#2}}
\providecommand{\bibinfo}[2]{#2}
\ifx\xfnm\relax \def\xfnm[#1]{\unskip,\space#1}\fi
\bibitem[{Anderson et~al.(2011)Anderson, Nielsen, Froehlich, DuBray, Druzgal,
  Cariello, Cooperrider, Zielinski, Ravichandran, Fletcher, Alexander, Bigler,
  Lange and Lainhart}]{anderson2011}
\bibinfo{author}{Anderson, J.S.}, \bibinfo{author}{Nielsen, J.A.},
  \bibinfo{author}{Froehlich, A.L.}, \bibinfo{author}{DuBray, M.B.},
  \bibinfo{author}{Druzgal, T.J.}, \bibinfo{author}{Cariello, A.N.},
  \bibinfo{author}{Cooperrider, J.R.}, \bibinfo{author}{Zielinski, B.A.},
  \bibinfo{author}{Ravichandran, C.}, \bibinfo{author}{Fletcher, P.T.},
  \bibinfo{author}{Alexander, A.L.}, \bibinfo{author}{Bigler, E.D.},
  \bibinfo{author}{Lange, N.}, \bibinfo{author}{Lainhart, J.E.},
  \bibinfo{year}{2011}.
\newblock \bibinfo{title}{{Functional connectivity magnetic resonance imaging
  classification of autism}}.
\newblock \bibinfo{journal}{Brain} \bibinfo{volume}{134},
  \bibinfo{pages}{3742--3754}.
\newblock
  \eprint{https://academic.oup.com/brain/article-pdf/134/12/3742/13795361/awr263.pdf}.
\bibitem[{Baio et~al.(2018)Baio, Wiggins, Christensen, Maenner, Daniels,
  Warren, Kurzius-Spencer, Zahorodny, Robinson, Rosenberg, White, Durkin, Imm,
  Nikolaou, Yeargin-Allsopp, Lee, Harrington, Lopez, Fitzgerald, Hewitt,
  Pettygrove, Constantino, Vehorn, Shenouda, Hall-Lande, Van, Naarden, Braun
  and Dowling}]{baio2018}
\bibinfo{author}{Baio, J.}, \bibinfo{author}{Wiggins, L.},
  \bibinfo{author}{Christensen, D.L.}, \bibinfo{author}{Maenner, M.J.},
  \bibinfo{author}{Daniels, J.}, \bibinfo{author}{Warren, Z.},
  \bibinfo{author}{Kurzius-Spencer, M.}, \bibinfo{author}{Zahorodny, W.},
  \bibinfo{author}{Robinson, C.}, \bibinfo{author}{Rosenberg},
  \bibinfo{author}{White, T.}, \bibinfo{author}{Durkin, M.S.},
  \bibinfo{author}{Imm, P.}, \bibinfo{author}{Nikolaou, L.},
  \bibinfo{author}{Yeargin-Allsopp, M.}, \bibinfo{author}{Lee, L.C.},
  \bibinfo{author}{Harrington, R.}, \bibinfo{author}{Lopez, M.},
  \bibinfo{author}{Fitzgerald, R.T.}, \bibinfo{author}{Hewitt, A.},
  \bibinfo{author}{Pettygrove, S.}, \bibinfo{author}{Constantino, J.N.},
  \bibinfo{author}{Vehorn, A.}, \bibinfo{author}{Shenouda, J.},
  \bibinfo{author}{Hall-Lande, J.}, \bibinfo{author}{Van, K.},
  \bibinfo{author}{Naarden}, \bibinfo{author}{Braun}, \bibinfo{author}{Dowling,
  N.F.}, \bibinfo{year}{2018}.
\newblock \bibinfo{title}{{Prevalence of Autism Spectrum Disorder Among
  Children Aged 8 Years — Autism and Developmental Disabilities Monitoring
  Network, 11 Sites, United States, 2014}}.
\newblock \bibinfo{journal}{MMWR. Surveillance Summaries} \bibinfo{volume}{67},
  \bibinfo{pages}{1--23}.
\bibitem[{Bassett and Sporns(2017)}]{Bassett:2017}
\bibinfo{author}{Bassett, D.S.}, \bibinfo{author}{Sporns, O.},
  \bibinfo{year}{2017}.
\newblock \bibinfo{title}{Network neuroscience}.
\newblock \bibinfo{journal}{Nature neuroscience} \bibinfo{volume}{20},
  \bibinfo{pages}{353}.
\bibitem[{Belcastro et~al.(2018)Belcastro, Poussin, Xiang, Giordano, Tripathi,
  Boda, Balci, Bilgen, Dhanda, Duan, Gong, Kumar, Romero, Sarac, Tarca, Wang,
  Yang, Yang, Zhang, Boué, Guarracino, Martin, Peitsch and
  Hoeng}]{belcastro2018}
\bibinfo{author}{Belcastro, V.}, \bibinfo{author}{Poussin, C.},
  \bibinfo{author}{Xiang, Y.}, \bibinfo{author}{Giordano, M.},
  \bibinfo{author}{Tripathi, K.P.}, \bibinfo{author}{Boda, A.},
  \bibinfo{author}{Balci, A.T.}, \bibinfo{author}{Bilgen, I.},
  \bibinfo{author}{Dhanda, S.K.}, \bibinfo{author}{Duan, Z.},
  \bibinfo{author}{Gong, X.}, \bibinfo{author}{Kumar, R.},
  \bibinfo{author}{Romero, R.}, \bibinfo{author}{Sarac, O.S.},
  \bibinfo{author}{Tarca, A.L.}, \bibinfo{author}{Wang, P.},
  \bibinfo{author}{Yang, H.}, \bibinfo{author}{Yang, W.},
  \bibinfo{author}{Zhang, C.}, \bibinfo{author}{Boué, S.},
  \bibinfo{author}{Guarracino, M.R.}, \bibinfo{author}{Martin, F.},
  \bibinfo{author}{Peitsch, M.C.}, \bibinfo{author}{Hoeng, J.},
  \bibinfo{year}{2018}.
\newblock \bibinfo{title}{The sbv improver systems toxicology computational
  challenge: Identification of human and species-independent blood response
  markers as predictors of smoking exposure and cessation status}.
\newblock \bibinfo{journal}{Computational Toxicology} \bibinfo{volume}{5},
  \bibinfo{pages}{38 -- 51}.
\bibitem[{Bergstra and Bengio(2012)}]{bergstra2012random}
\bibinfo{author}{Bergstra, J.}, \bibinfo{author}{Bengio, Y.},
  \bibinfo{year}{2012}.
\newblock \bibinfo{title}{Random search for hyper-parameter optimization}.
\newblock \bibinfo{journal}{Journal of machine learning research}
  \bibinfo{volume}{13}, \bibinfo{pages}{281--305}.
\bibitem[{Breiman(1996)}]{breiman1996bagging}
\bibinfo{author}{Breiman, L.}, \bibinfo{year}{1996}.
\newblock \bibinfo{title}{Bagging predictors}.
\newblock \bibinfo{journal}{Machine learning} \bibinfo{volume}{24},
  \bibinfo{pages}{123--140}.
\bibitem[{Breiman(1999)}]{breiman1999pasting}
\bibinfo{author}{Breiman, L.}, \bibinfo{year}{1999}.
\newblock \bibinfo{title}{Pasting small votes for classification in large
  databases and on-line}.
\newblock \bibinfo{journal}{Machine learning} \bibinfo{volume}{36},
  \bibinfo{pages}{85--103}.
\bibitem[{Breiman(2001)}]{breiman2001random}
\bibinfo{author}{Breiman, L.}, \bibinfo{year}{2001}.
\newblock \bibinfo{title}{Random forests}.
\newblock \bibinfo{journal}{Machine learning} \bibinfo{volume}{45},
  \bibinfo{pages}{5--32}.
\bibitem[{Breiman et~al.(1984)Breiman, Friedman, Stone and
  Olshen}]{breiman1984classification}
\bibinfo{author}{Breiman, L.}, \bibinfo{author}{Friedman, J.},
  \bibinfo{author}{Stone, C.J.}, \bibinfo{author}{Olshen, R.A.},
  \bibinfo{year}{1984}.
\newblock \bibinfo{title}{Classification and regression trees}.
\newblock \bibinfo{publisher}{CRC press}.
\bibitem[{Bron et~al.(2015)Bron, Smits, Van Der~Flier, Vrenken, Barkhof,
  Scheltens, Papma, Steketee, Orellana, Meijboom et~al.}]{bron2015standardized}
\bibinfo{author}{Bron, E.E.}, \bibinfo{author}{Smits, M.}, \bibinfo{author}{Van
  Der~Flier, W.M.}, \bibinfo{author}{Vrenken, H.}, \bibinfo{author}{Barkhof,
  F.}, \bibinfo{author}{Scheltens, P.}, \bibinfo{author}{Papma, J.M.},
  \bibinfo{author}{Steketee, R.M.}, \bibinfo{author}{Orellana, C.M.},
  \bibinfo{author}{Meijboom, R.}, et~al., \bibinfo{year}{2015}.
\newblock \bibinfo{title}{Standardized evaluation of algorithms for
  computer-aided diagnosis of dementia based on structural mri: the caddementia
  challenge}.
\newblock \bibinfo{journal}{NeuroImage} \bibinfo{volume}{111},
  \bibinfo{pages}{562--579}.
\bibitem[{Brown and Hamarneh(2016)}]{brown2016}
\bibinfo{author}{Brown, C.J.}, \bibinfo{author}{Hamarneh, G.},
  \bibinfo{year}{2016}.
\newblock \bibinfo{title}{Machine learning on human connectome data from mri}.
\newblock \eprint{arXiv:1611.08699}.
\bibitem[{Chang and Lin(2011)}]{chang2011libsvm}
\bibinfo{author}{Chang, C.C.}, \bibinfo{author}{Lin, C.J.},
  \bibinfo{year}{2011}.
\newblock \bibinfo{title}{Libsvm: A library for support vector machines}.
\newblock \bibinfo{journal}{ACM transactions on intelligent systems and
  technology (TIST)} \bibinfo{volume}{2}, \bibinfo{pages}{1--27}.
\bibitem[{Chen and Guestrin(2016)}]{chen2016xgboost}
\bibinfo{author}{Chen, T.}, \bibinfo{author}{Guestrin, C.},
  \bibinfo{year}{2016}.
\newblock \bibinfo{title}{Xgboost: A scalable tree boosting system}, in:
  \bibinfo{booktitle}{Proceedings of the 22nd acm sigkdd international
  conference on knowledge discovery and data mining}, pp.
  \bibinfo{pages}{785--794}.
\bibitem[{{Debnath} et~al.(2018){Debnath}, {Biswas}, {Ashik} and
  {Dash}}]{debnath2018}
\bibinfo{author}{{Debnath}, T.}, \bibinfo{author}{{Biswas}, T.},
  \bibinfo{author}{{Ashik}, M.H.}, \bibinfo{author}{{Dash}, S.},
  \bibinfo{year}{2018}.
\newblock \bibinfo{title}{Auto-encoder based nonlinear dimensionality reduction
  of ecg data and classification of cardiac arrhythmia groups using deep neural
  network}, in: \bibinfo{booktitle}{2018 4th International Conference on
  Electrical Engineering and Information Communication Technology (iCEEiCT)},
  pp. \bibinfo{pages}{27--31}.
\bibitem[{Dekhil et~al.(2019)Dekhil, Ali, El-Nakieb, Shalaby, Soliman, Switala,
  Mahmoud, Ghazal, Hajjdiab, Casanova, Elmaghraby, Keynton, El-Baz and
  Barnes}]{dekhil2019}
\bibinfo{author}{Dekhil, O.}, \bibinfo{author}{Ali, M.},
  \bibinfo{author}{El-Nakieb, Y.}, \bibinfo{author}{Shalaby, A.},
  \bibinfo{author}{Soliman, A.}, \bibinfo{author}{Switala, A.},
  \bibinfo{author}{Mahmoud, A.}, \bibinfo{author}{Ghazal, M.},
  \bibinfo{author}{Hajjdiab, H.}, \bibinfo{author}{Casanova, M.F.},
  \bibinfo{author}{Elmaghraby, A.}, \bibinfo{author}{Keynton, R.},
  \bibinfo{author}{El-Baz, A.}, \bibinfo{author}{Barnes, G.},
  \bibinfo{year}{2019}.
\newblock \bibinfo{title}{A personalized autism diagnosis cad system using a
  fusion of structural mri and resting-state functional mri data}.
\newblock \bibinfo{journal}{Frontiers in Psychiatry} \bibinfo{volume}{10},
  \bibinfo{pages}{392}.
\bibitem[{Di~Martino et~al.(2014)Di~Martino, Yan, Li, Denio, Castellanos,
  Alaerts, Anderson, Assaf, Bookheimer, Dapretto et~al.}]{Di-Martino:2014aa}
\bibinfo{author}{Di~Martino, A.}, \bibinfo{author}{Yan, C.G.},
  \bibinfo{author}{Li, Q.}, \bibinfo{author}{Denio, E.},
  \bibinfo{author}{Castellanos, F.X.}, \bibinfo{author}{Alaerts, K.},
  \bibinfo{author}{Anderson, J.S.}, \bibinfo{author}{Assaf, M.},
  \bibinfo{author}{Bookheimer, S.Y.}, \bibinfo{author}{Dapretto, M.}, et~al.,
  \bibinfo{year}{2014}.
\newblock \bibinfo{title}{The autism brain imaging data exchange: towards a
  large-scale evaluation of the intrinsic brain architecture in autism}.
\newblock \bibinfo{journal}{Molecular psychiatry} \bibinfo{volume}{19},
  \bibinfo{pages}{659}.
\bibitem[{Du et~al.(2018)Du, Fu and Calhoun}]{du2018classification}
\bibinfo{author}{Du, Y.}, \bibinfo{author}{Fu, Z.}, \bibinfo{author}{Calhoun,
  V.D.}, \bibinfo{year}{2018}.
\newblock \bibinfo{title}{Classification and prediction of brain disorders
  using functional connectivity: promising but challenging}.
\newblock \bibinfo{journal}{Frontiers in neuroscience} \bibinfo{volume}{12},
  \bibinfo{pages}{525}.
\bibitem[{Eslami et~al.(2019)Eslami, Mirjalili, Fong, Laird and
  Saeed}]{eslami2019}
\bibinfo{author}{Eslami, T.}, \bibinfo{author}{Mirjalili, V.},
  \bibinfo{author}{Fong, A.}, \bibinfo{author}{Laird, A.R.},
  \bibinfo{author}{Saeed, F.}, \bibinfo{year}{2019}.
\newblock \bibinfo{title}{Asd-diagnet: A hybrid learning approach for detection
  of autism spectrum disorder using fmri data}.
\newblock \bibinfo{journal}{Frontiers in Neuroinformatics}
  \bibinfo{volume}{13}, \bibinfo{pages}{70}.
\bibitem[{Fischl(2012)}]{Fischl:2012aa}
\bibinfo{author}{Fischl, B.}, \bibinfo{year}{2012}.
\newblock \bibinfo{title}{Freesurfer}.
\newblock \bibinfo{journal}{Neuroimage} \bibinfo{volume}{62},
  \bibinfo{pages}{774--781}.
\bibitem[{Fornito et~al.(2015)Fornito, Zalesky and Breakspear}]{Fornito:2015}
\bibinfo{author}{Fornito, A.}, \bibinfo{author}{Zalesky, A.},
  \bibinfo{author}{Breakspear, M.}, \bibinfo{year}{2015}.
\newblock \bibinfo{title}{The connectomics of brain disorders}.
\newblock \bibinfo{journal}{Nature Reviews Neuroscience} \bibinfo{volume}{16},
  \bibinfo{pages}{159--172}.
\bibitem[{Freund and Schapire(1995)}]{freund1995desicion}
\bibinfo{author}{Freund, Y.}, \bibinfo{author}{Schapire, R.E.},
  \bibinfo{year}{1995}.
\newblock \bibinfo{title}{A desicion-theoretic generalization of on-line
  learning and an application to boosting}, in: \bibinfo{booktitle}{European
  conference on computational learning theory},
  \bibinfo{organization}{Springer}. pp. \bibinfo{pages}{23--37}.
\bibitem[{Friedman(2001)}]{friedman2001greedy}
\bibinfo{author}{Friedman, J.H.}, \bibinfo{year}{2001}.
\newblock \bibinfo{title}{Greedy function approximation: a gradient boosting
  machine}.
\newblock \bibinfo{journal}{Annals of statistics} ,
  \bibinfo{pages}{1189--1232}.
\bibitem[{Georges et~al.(2020)Georges, Mhiri, Rekik, Initiative
  et~al.}]{georges2020}
\bibinfo{author}{Georges, N.}, \bibinfo{author}{Mhiri, I.},
  \bibinfo{author}{Rekik, I.}, \bibinfo{author}{Initiative, A.D.N.}, et~al.,
  \bibinfo{year}{2020}.
\newblock \bibinfo{title}{Identifying the best data-driven feature selection
  method for boosting reproducibility in classification tasks}.
\newblock \bibinfo{journal}{Pattern Recognition} , \bibinfo{pages}{107183}.
\bibitem[{Geschwind and Levitt(2007)}]{geschwind2007autism}
\bibinfo{author}{Geschwind, D.H.}, \bibinfo{author}{Levitt, P.},
  \bibinfo{year}{2007}.
\newblock \bibinfo{title}{Autism spectrum disorders: developmental
  disconnection syndromes}.
\newblock \bibinfo{journal}{Current opinion in neurobiology}
  \bibinfo{volume}{17}, \bibinfo{pages}{103--111}.
\bibitem[{GitHub(2017)}]{simlrpy}
\bibinfo{author}{GitHub, I.}, \bibinfo{year}{2017}.
\newblock \bibinfo{title}{Open source survey}.
\newblock \bibinfo{howpublished}{\url{https://github.com/bowang-lab/SIMLR_PY}}.
\bibitem[{Guyon et~al.(2002)Guyon, Weston, Barnhill and Vapnik}]{guyon2002gene}
\bibinfo{author}{Guyon, I.}, \bibinfo{author}{Weston, J.},
  \bibinfo{author}{Barnhill, S.}, \bibinfo{author}{Vapnik, V.},
  \bibinfo{year}{2002}.
\newblock \bibinfo{title}{Gene selection for cancer classification using
  support vector machines}.
\newblock \bibinfo{journal}{Machine learning} \bibinfo{volume}{46},
  \bibinfo{pages}{389--422}.
\bibitem[{Géron(2017)}]{géron2017hands-on}
\bibinfo{author}{Géron, A.}, \bibinfo{year}{2017}.
\newblock \bibinfo{title}{Hands-on machine learning with Scikit-Learn and
  TensorFlow : concepts, tools, and techniques to build intelligent systems}.
\newblock \bibinfo{publisher}{O'Reilly Media}, \bibinfo{address}{Sebastopol,
  CA}.
\bibitem[{Heinsfeld et~al.(2018)Heinsfeld, Franco, Craddock, Buchweitz and
  Meneguzzi}]{heinsfeld2018}
\bibinfo{author}{Heinsfeld, A.S.}, \bibinfo{author}{Franco, A.R.},
  \bibinfo{author}{Craddock, R.C.}, \bibinfo{author}{Buchweitz, A.},
  \bibinfo{author}{Meneguzzi, F.}, \bibinfo{year}{2018}.
\newblock \bibinfo{title}{Identification of autism spectrum disorder using deep
  learning and the abide dataset}.
\newblock \bibinfo{journal}{NeuroImage: Clinical} \bibinfo{volume}{17},
  \bibinfo{pages}{16 -- 23}.
\bibitem[{Hosmer~Jr et~al.(2013)Hosmer~Jr, Lemeshow and
  Sturdivant}]{hosmer2013applied}
\bibinfo{author}{Hosmer~Jr, D.W.}, \bibinfo{author}{Lemeshow, S.},
  \bibinfo{author}{Sturdivant, R.X.}, \bibinfo{year}{2013}.
\newblock \bibinfo{title}{Applied logistic regression}. volume
  \bibinfo{volume}{398}.
\newblock \bibinfo{publisher}{John Wiley \& Sons}.
\bibitem[{Hull et~al.(2017)Hull, Dokovna, Jacokes, Torgerson, Irimia and
  Van~Horn}]{hull2017resting}
\bibinfo{author}{Hull, J.V.}, \bibinfo{author}{Dokovna, L.B.},
  \bibinfo{author}{Jacokes, Z.J.}, \bibinfo{author}{Torgerson, C.M.},
  \bibinfo{author}{Irimia, A.}, \bibinfo{author}{Van~Horn, J.D.},
  \bibinfo{year}{2017}.
\newblock \bibinfo{title}{Resting-state functional connectivity in autism
  spectrum disorders: a review}.
\newblock \bibinfo{journal}{Frontiers in psychiatry} \bibinfo{volume}{7},
  \bibinfo{pages}{205}.
\bibitem[{Itahashi et~al.(2015)Itahashi, Yamada, Nakamura, Watanabe, Yamagata,
  Jimbo, Shioda, Kuroda, Toriizuka, Kato and Hashimoto}]{itahashi2015}
\bibinfo{author}{Itahashi, T.}, \bibinfo{author}{Yamada, T.},
  \bibinfo{author}{Nakamura, M.}, \bibinfo{author}{Watanabe, H.},
  \bibinfo{author}{Yamagata, B.}, \bibinfo{author}{Jimbo, D.},
  \bibinfo{author}{Shioda, S.}, \bibinfo{author}{Kuroda, M.},
  \bibinfo{author}{Toriizuka, K.}, \bibinfo{author}{Kato, N.},
  \bibinfo{author}{Hashimoto, R.}, \bibinfo{year}{2015}.
\newblock \bibinfo{title}{Linked alterations in gray and white matter
  morphology in adults with high-functioning autism spectrum disorder: A
  multimodal brain imaging study}.
\newblock \bibinfo{journal}{NeuroImage: Clinical} \bibinfo{volume}{7},
  \bibinfo{pages}{155 -- 169}.
\bibitem[{Janecek et~al.(2008)Janecek, Gansterer, Demel and
  Ecker}]{janecek2008}
\bibinfo{author}{Janecek, A.}, \bibinfo{author}{Gansterer, W.},
  \bibinfo{author}{Demel, M.}, \bibinfo{author}{Ecker, G.},
  \bibinfo{year}{2008}.
\newblock \bibinfo{title}{On the relationship between feature selection and
  classification accuracy}, in: \bibinfo{editor}{Saeys, Y.},
  \bibinfo{editor}{Liu, H.}, \bibinfo{editor}{Inza, I.},
  \bibinfo{editor}{Wehenkel, L.}, \bibinfo{editor}{de~Pee, Y.V.} (Eds.),
  \bibinfo{booktitle}{Proceedings of the Workshop on New Challenges for Feature
  Selection in Data Mining and Knowledge Discovery at ECML/PKDD 2008},
  \bibinfo{publisher}{PMLR}, \bibinfo{address}{Antwerp, Belgium}. pp.
  \bibinfo{pages}{90--105}.
\bibitem[{{Kaggle}()}]{Kaggle}
\bibinfo{author}{{Kaggle}}, .
\bibitem[{Kiarashinejad et~al.(2020)Kiarashinejad, Abdollahramezani and
  Adibi}]{kiarashinejad2020deep}
\bibinfo{author}{Kiarashinejad, Y.}, \bibinfo{author}{Abdollahramezani, S.},
  \bibinfo{author}{Adibi, A.}, \bibinfo{year}{2020}.
\newblock \bibinfo{title}{Deep learning approach based on dimensionality
  reduction for designing electromagnetic nanostructures}.
\newblock \bibinfo{journal}{npj Computational Materials} \bibinfo{volume}{6},
  \bibinfo{pages}{1--12}.
\bibitem[{Kittler et~al.(1998)Kittler, Hatef, Duin and
  Matas}]{kittler1998combining}
\bibinfo{author}{Kittler, J.}, \bibinfo{author}{Hatef, M.},
  \bibinfo{author}{Duin, R.P.}, \bibinfo{author}{Matas, J.},
  \bibinfo{year}{1998}.
\newblock \bibinfo{title}{On combining classifiers}.
\newblock \bibinfo{journal}{IEEE transactions on pattern analysis and machine
  intelligence} \bibinfo{volume}{20}, \bibinfo{pages}{226--239}.
\bibitem[{Lang et~al.(2016)Lang, Hancock and Singh}]{lang2016early}
\bibinfo{author}{Lang, R.}, \bibinfo{author}{Hancock, T.B.},
  \bibinfo{author}{Singh, N.N.}, \bibinfo{year}{2016}.
\newblock \bibinfo{title}{Early intervention for young children with autism
  spectrum disorder}.
\newblock \bibinfo{publisher}{Springer}.
\bibitem[{Lisowska et~al.(2019)Lisowska, Rekik, AbbVie, Foundation, Biotech,
  Bio-Clinica, Biogen, Company, CereSpir, Cogstate et~al.}]{lisowska2019joint}
\bibinfo{author}{Lisowska, A.}, \bibinfo{author}{Rekik, I.},
  \bibinfo{author}{AbbVie, A.A.}, \bibinfo{author}{Foundation, A.D.D.},
  \bibinfo{author}{Biotech, A.}, \bibinfo{author}{Bio-Clinica, I.},
  \bibinfo{author}{Biogen}, \bibinfo{author}{Company, B.M.S.},
  \bibinfo{author}{CereSpir, I.}, \bibinfo{author}{Cogstate}, et~al.,
  \bibinfo{year}{2019}.
\newblock \bibinfo{title}{Joint pairing and structured mapping of convolutional
  brain morphological multiplexes for early dementia diagnosis}.
\newblock \bibinfo{journal}{Brain connectivity} \bibinfo{volume}{9},
  \bibinfo{pages}{22--36}.
\bibitem[{Liu et~al.(2008)Liu, Ting and Zhou}]{liu2008isolation}
\bibinfo{author}{Liu, F.T.}, \bibinfo{author}{Ting, K.M.},
  \bibinfo{author}{Zhou, Z.H.}, \bibinfo{year}{2008}.
\newblock \bibinfo{title}{Isolation forest}, in: \bibinfo{booktitle}{2008
  Eighth IEEE International Conference on Data Mining},
  \bibinfo{organization}{IEEE}. pp. \bibinfo{pages}{413--422}.
\bibitem[{Mahjoub et~al.(2018)Mahjoub, Mahjoub and Rekik}]{mahjoub2018brain}
\bibinfo{author}{Mahjoub, I.}, \bibinfo{author}{Mahjoub, M.A.},
  \bibinfo{author}{Rekik, I.}, \bibinfo{year}{2018}.
\newblock \bibinfo{title}{Brain multiplexes reveal morphological connectional
  biomarkers fingerprinting late brain dementia states}.
\newblock \bibinfo{journal}{Scientific reports} \bibinfo{volume}{8},
  \bibinfo{pages}{1--14}.
\bibitem[{Marbach et~al.(2012)Marbach, Costello, K{\"u}ffner, Vega, Prill,
  Camacho, Allison, Aderhold, Bonneau, Chen et~al.}]{marbach2012wisdom}
\bibinfo{author}{Marbach, D.}, \bibinfo{author}{Costello, J.C.},
  \bibinfo{author}{K{\"u}ffner, R.}, \bibinfo{author}{Vega, N.M.},
  \bibinfo{author}{Prill, R.J.}, \bibinfo{author}{Camacho, D.M.},
  \bibinfo{author}{Allison, K.R.}, \bibinfo{author}{Aderhold, A.},
  \bibinfo{author}{Bonneau, R.}, \bibinfo{author}{Chen, Y.}, et~al.,
  \bibinfo{year}{2012}.
\newblock \bibinfo{title}{Wisdom of crowds for robust gene network inference}.
\newblock \bibinfo{journal}{Nature methods} \bibinfo{volume}{9},
  \bibinfo{pages}{796}.
\bibitem[{Morris and Rekik(2017)}]{morris2017}
\bibinfo{author}{Morris, C.}, \bibinfo{author}{Rekik, I.},
  \bibinfo{year}{2017}.
\newblock \bibinfo{title}{{Autism Spectrum Disorder Diagnosis Using Sparse
  Graph Embedding of Morphological Brain Networks}}, in:
  \bibinfo{editor}{Cardoso, M.J.}, \bibinfo{editor}{Arbel, T.},
  \bibinfo{editor}{Ferrante, E.}, \bibinfo{editor}{Pennec, X.},
  \bibinfo{editor}{Dalca, A.V.}, \bibinfo{editor}{Parisot, S.},
  \bibinfo{editor}{Joshi, S.}, \bibinfo{editor}{Batmanghelich, N.K.},
  \bibinfo{editor}{Sotiras, A.}, \bibinfo{editor}{Nielsen, M.},
  \bibinfo{editor}{Sabuncu, M.R.}, \bibinfo{editor}{Fletcher, T.},
  \bibinfo{editor}{Shen, L.}, \bibinfo{editor}{Durrleman, S.},
  \bibinfo{editor}{Sommer, S.} (Eds.), \bibinfo{booktitle}{Graphs in Biomedical
  Image Analysis, Computational Anatomy and Imaging Genetics},
  \bibinfo{publisher}{Springer International Publishing},
  \bibinfo{address}{Cham}. pp. \bibinfo{pages}{12--20}.
\bibitem[{Pappu and Pardalos(2014)}]{pappu2014high}
\bibinfo{author}{Pappu, V.}, \bibinfo{author}{Pardalos, P.M.},
  \bibinfo{year}{2014}.
\newblock \bibinfo{title}{High-dimensional data classification}, in:
  \bibinfo{booktitle}{Clusters, Orders, and Trees: Methods and Applications}.
  \bibinfo{publisher}{Springer}, pp. \bibinfo{pages}{119--150}.
\bibitem[{Pedregosa et~al.(2011)Pedregosa, Varoquaux, Gramfort, Michel,
  Thirion, Grisel, Blondel, Prettenhofer, Weiss, Dubourg, Vanderplas, Passos,
  Cournapeau, Brucher, Perrot and Duchesnay}]{scikit-learn}
\bibinfo{author}{Pedregosa, F.}, \bibinfo{author}{Varoquaux, G.},
  \bibinfo{author}{Gramfort, A.}, \bibinfo{author}{Michel, V.},
  \bibinfo{author}{Thirion, B.}, \bibinfo{author}{Grisel, O.},
  \bibinfo{author}{Blondel, M.}, \bibinfo{author}{Prettenhofer, P.},
  \bibinfo{author}{Weiss, R.}, \bibinfo{author}{Dubourg, V.},
  \bibinfo{author}{Vanderplas, J.}, \bibinfo{author}{Passos, A.},
  \bibinfo{author}{Cournapeau, D.}, \bibinfo{author}{Brucher, M.},
  \bibinfo{author}{Perrot, M.}, \bibinfo{author}{Duchesnay, E.},
  \bibinfo{year}{2011}.
\newblock \bibinfo{title}{Scikit-learn: Machine learning in {P}ython}.
\newblock \bibinfo{journal}{Journal of Machine Learning Research}
  \bibinfo{volume}{12}, \bibinfo{pages}{2825--2830}.
\bibitem[{Peng et~al.(2005)Peng, Long and Ding}]{article}
\bibinfo{author}{Peng, H.}, \bibinfo{author}{Long, F.}, \bibinfo{author}{Ding,
  C.}, \bibinfo{year}{2005}.
\newblock \bibinfo{title}{Feature selection based on mutual information:
  Criteria of max-dependency,max-relevance, and min-redundancy}.
\newblock \bibinfo{journal}{IEEE transactions on pattern analysis and machine
  intelligence} \bibinfo{volume}{27}, \bibinfo{pages}{1226--38}.
\bibitem[{Peterson(2009)}]{Peterson:2009}
\bibinfo{author}{Peterson, L.E.}, \bibinfo{year}{2009}.
\newblock \bibinfo{title}{{K}-nearest neighbor}.
\newblock \bibinfo{journal}{Scholarpedia} \bibinfo{volume}{4},
  \bibinfo{pages}{1883}.
\newblock \bibinfo{note}{Revision \#137311}.
\bibitem[{Postema et~al.(2019)Postema, Van~Rooij, Anagnostou, Arango, Auzias,
  Behrmann, Busatto, Calderoni, Calvo, Daly, Deruelle, Di~Martino, Dinstein,
  Duran, Durston, Ecker, Ehrlich, Fair, Fedor and Francks}]{postema2019}
\bibinfo{author}{Postema, M.}, \bibinfo{author}{Van~Rooij, D.},
  \bibinfo{author}{Anagnostou, E.}, \bibinfo{author}{Arango, C.},
  \bibinfo{author}{Auzias, G.}, \bibinfo{author}{Behrmann, M.},
  \bibinfo{author}{Busatto, G.}, \bibinfo{author}{Calderoni, S.},
  \bibinfo{author}{Calvo, R.}, \bibinfo{author}{Daly, E.},
  \bibinfo{author}{Deruelle, C.}, \bibinfo{author}{Di~Martino, A.},
  \bibinfo{author}{Dinstein, I.}, \bibinfo{author}{Duran, F.},
  \bibinfo{author}{Durston, S.}, \bibinfo{author}{Ecker, C.},
  \bibinfo{author}{Ehrlich, S.}, \bibinfo{author}{Fair, D.},
  \bibinfo{author}{Fedor, J.}, \bibinfo{author}{Francks, C.},
  \bibinfo{year}{2019}.
\newblock \bibinfo{title}{Altered structural brain asymmetry in autism spectrum
  disorder in a study of 54 datasets}.
\newblock \bibinfo{journal}{Nature Communications} \bibinfo{volume}{10}.
\bibitem[{Raeper et~al.(2018)Raeper, Lisowska and
  Rekik}]{raeper2018cooperative}
\bibinfo{author}{Raeper, R.}, \bibinfo{author}{Lisowska, A.},
  \bibinfo{author}{Rekik, I.}, \bibinfo{year}{2018}.
\newblock \bibinfo{title}{Cooperative correlational and discriminative ensemble
  classifier learning for early dementia diagnosis using morphological brain
  multiplexes}.
\newblock \bibinfo{journal}{IEEE Access} \bibinfo{volume}{6},
  \bibinfo{pages}{43830--43839}.
\bibitem[{Randall and Williams(2018)}]{randall2018}
\bibinfo{author}{Randall, M, E.K.S.A.S.R.H.L.L.N.S.K.W.S.},
  \bibinfo{author}{Williams, K.}, \bibinfo{year}{2018}.
\newblock \bibinfo{title}{Diagnostic tests for autism spectrum disorder (asd)
  in preschool children}.
\newblock \bibinfo{journal}{Cochrane Database of Systematic Reviews} .
\bibitem[{Raschka(2014a)}]{raschka2014feature}
\bibinfo{author}{Raschka, S.}, \bibinfo{year}{2014}a.
\newblock \bibinfo{title}{About feature scaling and normalization}.
\newblock \bibinfo{journal}{Sebastian Racha. Disques, nd Web. Dec} .
\bibitem[{Raschka(2014b)}]{raschka2014linear}
\bibinfo{author}{Raschka, S.}, \bibinfo{year}{2014}b.
\newblock \bibinfo{title}{Linear discriminant analysis bit by bit}.
\newblock \bibinfo{journal}{Disponible en: sebastianraschka.
  com/Articles/2014\_python\_lda. html} .
\bibitem[{Saez-Rodriguez et~al.(2016)Saez-Rodriguez, Costello, Friend, Kellen,
  Mangravite, Meyer, Norman and Stolovitzky}]{rodriguez2016}
\bibinfo{author}{Saez-Rodriguez, J.}, \bibinfo{author}{Costello, J.C.},
  \bibinfo{author}{Friend, S.H.}, \bibinfo{author}{Kellen, M.R.},
  \bibinfo{author}{Mangravite, L.}, \bibinfo{author}{Meyer, P.},
  \bibinfo{author}{Norman, T.}, \bibinfo{author}{Stolovitzky, G.},
  \bibinfo{year}{2016}.
\newblock \bibinfo{title}{{Crowdsourcing biomedical research: Leveraging
  communities as innovation engines}}.
\bibitem[{Soussia and Rekik(2017)}]{soussia2017}
\bibinfo{author}{Soussia, M.}, \bibinfo{author}{Rekik, I.},
  \bibinfo{year}{2017}.
\newblock \bibinfo{title}{High-order connectomic manifold learning for autistic
  brain state identification}, in: \bibinfo{booktitle}{International Workshop
  on Connectomics in Neuroimaging}, \bibinfo{organization}{Springer}. pp.
  \bibinfo{pages}{51--59}.
\bibitem[{Soussia and Rekik(2018)}]{soussia2018}
\bibinfo{author}{Soussia, M.}, \bibinfo{author}{Rekik, I.},
  \bibinfo{year}{2018}.
\newblock \bibinfo{title}{Unsupervised manifold learning using high-order
  morphological brain networks derived from t1-w mri for autism diagnosis}.
\newblock \bibinfo{journal}{Frontiers in Neuroinformatics}
  \bibinfo{volume}{12}, \bibinfo{pages}{70}.
\bibitem[{Tenenbaum et~al.(2000)Tenenbaum, Silva and Langford}]{tenenbaum2000}
\bibinfo{author}{Tenenbaum, J.B.}, \bibinfo{author}{Silva, V.d.},
  \bibinfo{author}{Langford, J.C.}, \bibinfo{year}{2000}.
\newblock \bibinfo{title}{A global geometric framework for nonlinear
  dimensionality reduction}.
\newblock \bibinfo{journal}{Science} \bibinfo{volume}{290},
  \bibinfo{pages}{2319--2323}.
\newblock
  \eprint{https://science.sciencemag.org/content/290/5500/2319.full.pdf}.
\bibitem[{Wang et~al.(2017)Wang, Zhu, Pierson, Ramazzotti and
  Batzoglou}]{wang2017visualization}
\bibinfo{author}{Wang, B.}, \bibinfo{author}{Zhu, J.},
  \bibinfo{author}{Pierson, E.}, \bibinfo{author}{Ramazzotti, D.},
  \bibinfo{author}{Batzoglou, S.}, \bibinfo{year}{2017}.
\newblock \bibinfo{title}{Visualization and analysis of single-cell rna-seq
  data by kernel-based similarity learning}.
\newblock \bibinfo{journal}{Nature methods} \bibinfo{volume}{14},
  \bibinfo{pages}{414}.
\bibitem[{Yang et~al.(2016)Yang, Beam, Pelphrey, Abdullahi and Jou}]{yang2016}
\bibinfo{author}{Yang, D.Y.J.}, \bibinfo{author}{Beam, D.},
  \bibinfo{author}{Pelphrey, K.A.}, \bibinfo{author}{Abdullahi, S.},
  \bibinfo{author}{Jou, R.J.}, \bibinfo{year}{2016}.
\newblock \bibinfo{title}{Cortical morphological markers in children with
  autism: a structural magnetic resonance imaging study of thickness, area,
  volume, and gyrification}.
\newblock \bibinfo{journal}{Molecular autism} \bibinfo{volume}{7},
  \bibinfo{pages}{11}.
\bibitem[{Zhang and Ma(2012)}]{zhang2012ensemble}
\bibinfo{author}{Zhang, C.}, \bibinfo{author}{Ma, Y.}, \bibinfo{year}{2012}.
\newblock \bibinfo{title}{Ensemble machine learning: methods and applications}.
\newblock \bibinfo{publisher}{Springer}.
\bibitem[{Zhang(2015)}]{zhang2014}
\bibinfo{author}{Zhang, X.L.}, \bibinfo{year}{2015}.
\newblock \bibinfo{title}{Nonlinear dimensionality reduction of data by deep
  distributed random samplings}, in: \bibinfo{editor}{Phung, D.},
  \bibinfo{editor}{Li, H.} (Eds.), \bibinfo{booktitle}{Proceedings of the Sixth
  Asian Conference on Machine Learning}, \bibinfo{publisher}{PMLR},
  \bibinfo{address}{Nha Trang City, Vietnam}. pp. \bibinfo{pages}{221--233}.
\bibitem[{Zhao et~al.(2018)Zhao, Zhang, Rekik, An and Shen}]{zhao2018}
\bibinfo{author}{Zhao, F.}, \bibinfo{author}{Zhang, H.},
  \bibinfo{author}{Rekik, I.}, \bibinfo{author}{An, Z.}, \bibinfo{author}{Shen,
  D.}, \bibinfo{year}{2018}.
\newblock \bibinfo{title}{Diagnosis of autism spectrum disorders using
  multi-level high-order functional networks derived from resting-state
  functional mri}.
\newblock \bibinfo{journal}{Frontiers in Human Neuroscience}
  \bibinfo{volume}{12}, \bibinfo{pages}{184}.
\bibitem[{Zwaigenbaum et~al.(2015)Zwaigenbaum, Bauman, Stone, Yirmiya, Estes,
  Hansen, McPartland, Natowicz, Choueiri, Fein, Kasari, Pierce, Buie, Carter,
  Davis, Granpeesheh, Mailloux, Newschaffer, Robins, Roley, Wagner and
  Wetherby}]{zwaigenbaum2015}
\bibinfo{author}{Zwaigenbaum, L.}, \bibinfo{author}{Bauman, M.L.},
  \bibinfo{author}{Stone, W.L.}, \bibinfo{author}{Yirmiya, N.},
  \bibinfo{author}{Estes, A.}, \bibinfo{author}{Hansen, R.L.},
  \bibinfo{author}{McPartland, J.C.}, \bibinfo{author}{Natowicz, M.R.},
  \bibinfo{author}{Choueiri, R.}, \bibinfo{author}{Fein, D.},
  \bibinfo{author}{Kasari, C.}, \bibinfo{author}{Pierce, K.},
  \bibinfo{author}{Buie, T.}, \bibinfo{author}{Carter, A.},
  \bibinfo{author}{Davis, P.A.}, \bibinfo{author}{Granpeesheh, D.},
  \bibinfo{author}{Mailloux, Z.}, \bibinfo{author}{Newschaffer, C.},
  \bibinfo{author}{Robins, D.}, \bibinfo{author}{Roley, S.S.},
  \bibinfo{author}{Wagner, S.}, \bibinfo{author}{Wetherby, A.},
  \bibinfo{year}{2015}.
\newblock \bibinfo{title}{{Early identification of autism spectrum disorder:
  Recommendations for practice and research}}, in:
  \bibinfo{booktitle}{Pediatrics}, \bibinfo{publisher}{American Academy of
  Pediatrics}. pp. \bibinfo{pages}{S10--S40}.

\end{thebibliography}
\bibliographystyle{model2-names}

\end{document}